\documentclass{article}

\usepackage[table]{xcolor}
\usepackage{array}
\usepackage{tabularx}
\usepackage{multirow}

\usepackage{microtype}
\usepackage{graphicx}
\usepackage{booktabs} 

\usepackage{hyperref}

\usepackage[accepted]{icml2025}

\usepackage{amsmath}
\usepackage{amssymb}
\usepackage{mathtools}
\usepackage{amsthm}

\usepackage[capitalize,noabbrev]{cleveref}

\usepackage{subcaption}
\usepackage{pgfplots}
\pgfplotsset{compat=1.17}
\usepgfplotslibrary{groupplots}
\usepgfplotslibrary{colormaps,patchplots}

\theoremstyle{plain}
\newtheorem{theorem}{Theorem}[section]

\theoremstyle{definition}

\theoremstyle{remark}

\usepackage[textsize=tiny]{todonotes}

\icmltitlerunning{
CURing Large Models: Compression via CUR Decomposition
}

\begin{document}

\twocolumn[
\icmltitle{
CURing Large Models: Compression via CUR Decomposition
}




\begin{icmlauthorlist}
\icmlauthor{Sanghyeon Park}{sch}
\icmlauthor{Soo-Mook Moon}{sch}
\end{icmlauthorlist}

\icmlaffiliation{sch}{Department of Electrical and Computer Engineering, Seoul National University, Seoul, Republic of Korea}

\icmlcorrespondingauthor{Soo-Mook Moon}{smoon@snu.ac.kr}

\icmlkeywords{Machine Learning, ICML}

\begin{center}
    $^1$Department of Electrical and Computer Engineering, Seoul National University, Seoul, Republic of Korea \\
    \texttt{\{lukepark, smoon\}@snu.ac.kr}
\end{center}

\vskip 0.3in
]




\newcommand{\rulesep}{\unskip\ \vrule\ }

\begin{abstract}
    Large deep learning models have achieved remarkable success but are resource-intensive, posing challenges such as memory usage.
We introduce CURing, a novel model compression method based on CUR matrix decomposition, which approximates weight matrices as the product of selected columns (\( C \)) and rows (\( R \)), and a small linking matrix (\( U \)).
We apply this decomposition to weights chosen based on the combined influence of their magnitudes and activations.
By identifying and retaining informative rows and columns, CURing significantly reduces model size with minimal performance loss.
For example, it reduces Llama3.1-8B’s parameters to 7.32B (\(-9\text{\%}\)) in just 129 seconds, over 20 times faster than prior compression methods.
%

\end{abstract}

\section{Introduction} \label{section:introduction}

\begin{figure*}[!t]
    \footnotesize
    \centering
    \begin{subfigure}{0.25\textwidth}
        \centering
        \includegraphics[width=\linewidth]{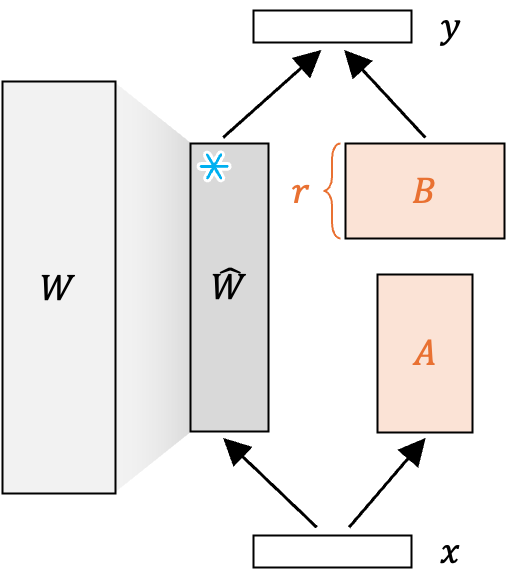}
        \caption{
            \textbf{Compression + LoRA}\\
            \textcolor{orange}{(e.g., $r = 8$)}\\
            \smallskip
            $y = x\widehat{W} + x{AB}$
        }
        \label{fig:subfig1}
    \end{subfigure}
    \hspace{12pt}
    \rulesep
    \hspace{12pt}
    \begin{subfigure}{0.25\textwidth}
        \centering
        \includegraphics[width=\linewidth]{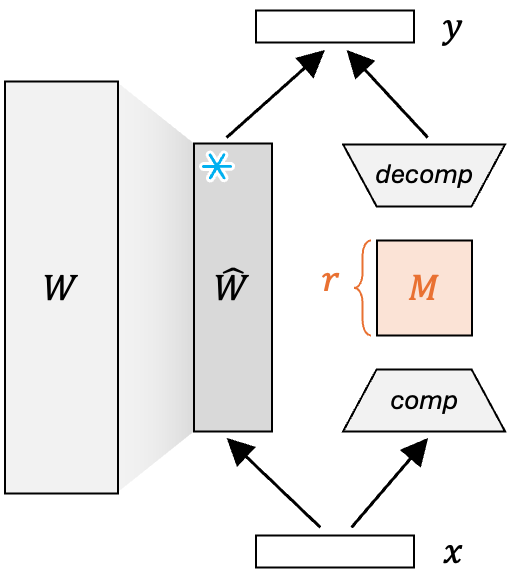}
        \caption{
            \textbf{Compression + MoRA}\\
            \textcolor{orange}{(e.g., $r = 256$)}\\
            \smallskip
            $y = x\widehat{W} + f_{decomp}(f_{comp}(x) {M})$
        }
        \label{fig:subfig2}
    \end{subfigure}
    \hspace{12pt}
    \rulesep
    \hspace{12pt}
    \begin{subfigure}{0.25\textwidth}
        \centering
        \includegraphics[width=\linewidth]{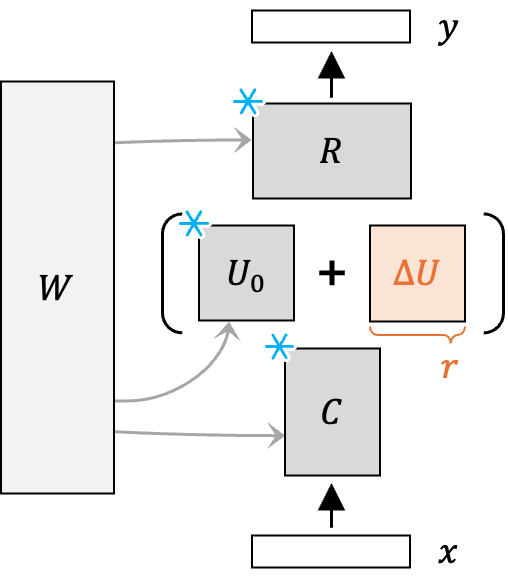}
        \caption{
            \textbf{CURing}\\
            \textcolor{orange}{(e.g., $r = 256$)}\\
            \smallskip
            $y = x({C} ({U}_0 + \Delta {U}) {R})$
        }
        \label{fig:subfig3}
    \end{subfigure}
    \caption{
        Comparison of compression-and-adaptation methods: LoRA, MoRA, and our proposed \textit{CURing}.
        Trainable parameters are in red, with \(r\) denoting rank.
        MoRA and CURing can use a larger \(r\) than LoRA without losing parameter efficiency.
        %
        Figures~\ref{fig:subfig1} and \ref{fig:subfig2} use a compressed model \(\widehat{W}\) (e.g., from pruning)
        with accuracy recovered by retraining low-rank matrices.
        However,
        CURing (Figure~\ref{fig:subfig3}) avoids retraining by using the low-parameter approximation \({W} \approx {C} {U}_0 {R}\).
        For further healing, we simply add a trainable matrix \(\Delta {U}\) to \({U}_0\),
        without incurring additional inference overhead.
    }
    \label{fig:adaptations}
\end{figure*}

The rapid advancement of deep learning has led to the development of increasingly large models that have achieved remarkable success across various domains~\cite{achiam2023gpt,dubey2024llama,liu2024deepseek}.
These models, while powerful, come with substantial memory requirements, making them challenging to deploy in resource-constrained environments.
In many practical applications, there is a critical need for models that are both accurate and efficient.

One approach to bridge this gap posed by compressing large models is the use of pruning then Parameter-Efficient Fine-Tuning (PEFT) as a form of healing~\cite{gromov2024unreasonable}.
In this context, models are first pruned to reduce their size by eliminating less significant parameters.
The pruning process, however, can lead to a loss of precision and degrade the model's performance.
Therefore, PEFT methods are then employed to retrain the model, efficiently healing its performance.
This combination enables the development of compact models that maintain high levels of accuracy where resources are limited.
However, even with the use of PEFT, retraining still requires considerable computational resources and a substantial amount of time.

Matrix decomposition is a promising approach for model compression, reducing neural network size while preserving key information.
By approximating the original matrix under low-rank conditions, it effectively minimizes storage and memory requirements without requiring retraining~\cite{chee2022model,flynn2024stat}.
However, highly information-preserving decompositions are often computationally expensive, both in factorization and in selecting elements to prune, making them less practical for large-scale models.
Additionally, decomposition can disrupt the original characteristics of weight matrices, such as explainability, as the factorized components consist of entirely new parameters distinct from the original matrix.

To address the challenges of model compressing—retraining overhead, massive process time, and losing characteristics—, we propose \textit{CURing}, a novel model compression technique based on CUR matrix decomposition.
By leveraging CUR decomposition's strong original-approximation feature, CURing inherently heals the damage caused by compression.
Unlike the (structural) pruning methods, CURing preserves the input/output dimensions, avoiding structural changes, while reducing the number of parameters by decomposing the original weight matrix \( W \) into the low-rank matrices \( C \), \( U \), and \( R \).
Furthermore, by adding a square matrix \( \Delta U \) to the linking matrix \( U_0 \gets U \) derived from CUR decomposition, CURing itself functions as a PEFT.
%
This allows for further parameter-efficient healing, however, updates are constrained to subspaces represented by \(C\) and \(R\) here, mitigating the forgetting that can occur during retraining.
This marks a significant distinction from using other PEFT methods.
In addition, since \( \Delta U \) is a square matrix, it has maximum expressiveness within the same rank. This enables CURing to be the best effective fine-tuning method like MoRA, even though its updates are constrained by the subspace.
Figure~\ref{fig:adaptations} provides a comprehensive visualization of CURing compared to LoRA~\cite{hu2021lora} and MoRA~\cite{jiang2024mora}.


In summary, the key contributions of this paper are:

\begin{itemize}

\item We introduce a novel neural network compression technique based on CUR decomposition, fast and effectively reducing model size while maintaining performance.
We demonstrate that our approach allows for automatic healing without retraining.


\item We show that CURing is also a parameter-efficient fine-tuning method itself, allowing a relatively high rank value within the constraint of same trainable parameters, and therefore enabling high-informative adaptation.
Furthermore, by constraining updates to the subspace, it mitigates forgetting during retraining.

\end{itemize}


\section{Related Work} \label{section:related-work}

\subsection{Pruning} \label{subsection:pruning}

Pruning reduces neural network size by removing or zeroizing significant weights or neurons.
\textit{Layer-wise Pruning} removes specific layers to improve efficiency.
In a recent study~\cite{gromov2024unreasonable}, similar layers were identified and removed by measuring angular distances between layer outputs in large GPT-style models, excluding the last layer.
After the pruning, fine-tuning using LoRA~\cite{hu2021lora} compensated for performance loss.
%
%
Another study~\cite{jha2024justchopembarrassinglysimple} explored selective removal of layers from decoder-based language models while keeping the first and last layers to preserve performance.
In other work, measuring the persistence of topological features in each layer led to the removal of layers when adjacent layers showed high similarity~\cite{gardinazzi2024persistent}.
\textit{Attention Pruning} removes unnecessary attention heads.
It was shown that only some attention heads in multi-head attention are important, and others can be removed without affecting performance~\cite{voita2019analyzing, michel2019sixteen}.
The recent research shows that making feed-forward network and query, key, value matrices sparse with most elements zero is possible with minimal performance degradation~\cite{jaszczur2021sparse}.
These studies suggest that, in LLMs, some layers and weights of attention can be replaced with low-rank approximations, supporting CURing's approach.

We can use additional information for better pruning.
The Fisher information matrix was used to perform precise pruning based on parameter influence on output distributions~\cite{van2023llm}, then LoRA was used to correct distortions from pruning.
%
%
WANDA~\cite{sun2023simple}, a method using input feature activations along with weight magnitudes for pruning, was proposed, allowing immediate use without retraining.
%
%

\subsection{Model Compression} \label{subsection:compression}

{Low-Rank approximations} have been widely used for model compression.
Self-attention matrices inherently possess low-dimensional characteristics, demonstrated via performance and Singular Value Decomposition (SVD) analysis~\cite{wang2019structured, wang2020linformer}.
Compression was also performed by lowering rank via SVD-based matrix factorization~\cite{wang2024bitstack, mao2020ladabert}.
%
%
%
Low-rank approximation through SVD in transformer FFN layers showed lower loss when pruning later (near-output) layers~\cite{sharma2023truth}.
Further, SliceGPT~\cite{ashkboos2024slicegpt} employs the Principal Component Analysis (PCA) technique to compress the weight matrices meticulously.

\textit{Interpolative Decomposition (ID)} has been used for compression, maintaining performance without extensive retraining by preserving original weight matrix information~\cite{chee2022model, flynn2024stat}.
%
%
The recent study, STAT~\cite{flynn2024stat}, uses QR decomposition to identify and remove less important parts, then generating correction matrices to minimize damage and maintain structure, eliminating the need for retraining.
CURing, like STAT, inherently possesses a correction effect (so no need to retrain) but achieves this faster in a single decomposition step without structural considerations, saving significant time (hours vs.\ minutes).  
While ID offers similar interpretability, CUR decomposition quickly obtains decomposed matrices~\cite{du2023matrix}.

\subsection{Parameter-Efficient Fine-Tuning} \label{subsection:peft}

Parameter-Efficient Fine-Tuning (PEFT) updates only a small number of parameters to adapt models efficiently.
{LoRA}~\cite{hu2021lora}, as in Figure~\ref{fig:subfig1}, learns two additional low-rank matrices during fine-tuning.
However, asymmetric low-rank matrices may have limitations due to low expressive power.
To overcome this, {MoRA}~\cite{jiang2024mora} uses square matrices for high-rank expressiveness with the same parameter efficiency, employing human-defined non-parameterized operators (\textit{comp}, \textit{decomp}) to compress and expand dimensions (Figure~\ref{fig:subfig2}).
%
CURing enables parameter-efficient high-rank updates via a trainable square matrix \( U \), achieving maximum rank for the same number of parameters.
By interpreting \( U \) as \( U_0 + \Delta U \), where initially \(U_0 \gets U\) and \(\Delta U \gets 0_{r \times r}\),
CURing can be seen as similar to MoRA, but differs in that it does not rely on human-defined modules.
Instead, \(\Delta U\) are constrained by the subspace defined by \(C\) and \(R\), enabling safe retraining (healing) without significant forgetting.

{CURLoRA}~\cite{fawi2024curlora} introduced CUR decomposition into LoRA to address catastrophic forgetting in PEFT.
Instead of LoRA's low-rank matrices, CURLoRA uses \( C \), \( U \), and \( R \), then fine-tuning only \( U \).
By sampling less important features for \( C \) and \( R \), it provides implicit regularization to prevent drastic changes when learning new tasks.
While CURing also uses \( \Delta U \) as a trainable parameter, it is fundamentally a model compression method, not just a PEFT technique.
Instead of sampling less important features, CURing captures the most important rows and columns to approximate the original matrix effectively.
The main focus is healing to mimic the original model's performance by updating \( U \), rather than learning new tasks.
%
For adapting CURing-compressed models to new tasks, PEFT methods like LoRA, MoRA, or CURLoRA can be used.
%
%
%

\subsection{Knowledge Distillation} \label{subsection:kd}

\textit{Knowledge Distillation} transfers knowledge from a large model to a smaller one~\cite{hinton2015distilling}.
%
%
Layer-wise differences between student and teacher models were expressed as mean squared error (MSE) loss for training~\cite{xia2022structured}.
Models were able to be compressed by training with block-specific losses~\cite{muralidharan2024compact}.
Our proposed CURing shows sufficient performance without retraining but employs distillation with the original model for additional healing.
%
Distillation on the C4 dataset compensates for loss caused by low-rank decomposition.
Although conducted solely on C4, the performance recovery is task-agnostic; experiments demonstrate strong recovery on multiple datasets including Wikitext, BoolQ, and MMLU.

\section{CUR Matrix Decomposition} \label{section:cur}

Matrix decomposition techniques are widely used for dimensionality reduction, data compression, and efficient computations~\cite{hamm2021perturbations, mahoney2009cur}.
A promising application is in compressing neural network models by approximating their weight matrices with low-rank representations.
The assumption is that model's core information which can be represented within lower rank exist; under this assumption, layers and components that have less impact (present less changes) can have their rank reduced.

%
\textit{CUR decomposition}
approximates an original matrix \( W \in \mathbb{R}^{m \times n} \) as:
\begin{equation}
    \notag
    W \approx C U R,
    \label{eq:cur}
\end{equation}
where
\( C = W\left[:, \mathbf{q}\right] \in \mathbb{R}^{m \times r} \) consists of selected columns from \( W \),
\( R = W\left[\mathbf{p}, :\right] \in \mathbb{R}^{r \times n} \) consists of selected rows from \( W \),
and
\( U \in \mathbb{R}^{r \times r} \) is a small square matrix capturing interactions between these rows and columns.
The integer vectors \( \mathbf{p}, \mathbf{q}\ \in \mathbb{N}^{r} \) are \( r \)-distinct selected indices.
CUR decomposition can approximate the original matrix well by properly selecting rows and columns, based on their importance (e.g., \( \ell_2 \)-norms~\cite{drineas2006fast2, drineas2006fast3} or leverage scores~\cite{mahoney2009cur, drineas2008relative}).

Once the matrices \( C \) and \( R \) are obtained, the core matrix \( U \) is computed:
\begin{equation}
    U = C^{\dagger} W R^{\dagger},
    \label{eq:U}
\end{equation}
where \( C^{\dagger} \) and \( R^{\dagger} \) are the pseudoinverses of \( C \) and \( R \), respectively~\cite{moore1920reciprocal}.
Computing \( U \) using pseudoinverses is optimal with respect to the Frobenius norm~\cite{stewart1999four}.


\subsection{DEIM-CUR} \label{subsection:deim-cur}

In CUR, various methods exist for sampling rows and columns efficiently.
Algorithms using random sampling probabilities based on norms or leverage scores allow for fast approximation~\cite{wang2012scalable, voronin2017efficient}, and even random selections yield satisfactory performance~\cite{boutsidis2014optimal, drineas2006fast2, drineas2006fast3, mahoney2009cur}.
However, these methods often require selecting more rows and columns than the target rank \( r \) to achieve bounded error performance.

Building on the \textit{Discrete Empirical Interpolation Method} (DEIM) selection algorithm~\cite{chaturantabut2010nonlinear, barrault2004empirical},
the DEIM-CUR decomposition~\cite{sorensen2016deim} offers a deterministic approach by selecting exactly \( r \) rows and \( r \) columns corresponding to the rank \( r \), leading to more accurate approximations under the constraint of limited selected rows and columns~\cite{hamm2020stability}.
Since the main purpose of this study is compression to reduce memory usage, adopting DEIM-CUR is more appropriate compared to other methods that require much more rows and columns.


The DEIM-CUR factorization provides a strong approximation of a matrix \( W \in \mathbb{R}^{m \times n} \) with a bounded error.
According to the studies~\cite{sorensen2016deim,drmac2016new}, the DEIM-CUR approximation is bounded within a factor of \( (\eta_p + \eta_q) \) relative to the error of the optimal rank-\(r\) solution (\( \sigma_{r+1} \))~\cite{eckart1936approximation}.
\begin{theorem}
    Let \( W \in \mathbb{R}^{m \times n} \) and \( 1 \le r \le \min(m, n) \).
    The rank-\(r\) singular value decomposition of \( W \) is expressed as \( W \approx P \Sigma Q^T \), where 
    \( P \in \mathbb{R}^{m \times r} \) and 
    \( Q \in \mathbb{R}^{n \times r} \) consist of the leading \( r \) left and right singular vectors, respectively.
    %
    Suppose the integer vectors \( \mathbf{p}, \mathbf{q} \in \mathbb{N}^r \) contain \( r \)-distinct indices selected using the DEIM algorithm from \( P \) and \( Q \), respectively (\( \mathbf{p} = \text{DEIM}(P) \) and \( \mathbf{q} = \text{DEIM}(Q) \)).
    The DEIM-CUR factorization defines the matrices
    \( C = W\left[:, \mathbf{q}\right] \in \mathbb{R}^{m \times r} \),
    \( R = W\left[\mathbf{p}, :\right] \in \mathbb{R}^{r \times n} \), and
    \( U = C^{\dagger} W R^{\dagger} \in \mathbb{R}^{r \times r} \).
    The error bound of the DEIM-CUR factorization is:
    \[
        \| W - CUR \|_2 \le (\eta_p + \eta_q) \sigma_{r+1},
    \]
    where \( \sigma_{r+1} \) is the first neglected singular value of \( W \), and the finite error constants are defined as
    \( \eta_p \equiv \| (P\left[\mathbf{p}, :\right])^{-1} \|_2 \) and
    \(   \eta_q \equiv \| (Q\left[:, \mathbf{q}\right])^{-1} \|_2 \).
    \label{theorem:deim}
\end{theorem}
%
%
%
%
Following the recent research~\cite{drmac2016new}, the DEIM-CUR factorization provides an improved and interpretable error bound given by:
\[
    \eta_p < \sqrt{\frac{mr}{3}}2^r, \quad \eta_q < \sqrt{\frac{nr}{3}}2^r.
\]

\subsection{Parameter Reduction} \label{subsection:parameter-reduction}

CUR decomposition effectively reduces the number of parameters in the model.
Specifically, the total number of parameters in \( C \), \( U \), and \( R \) is smaller than in the original matrix \( W \) when the condition \( mn > mr + r^2 + rn \) is met, where \( r \) represents the rank.
Specifically, we use \( r \ll \min(m, n) \) such that:
\begin{equation}
    r \gets \min{\left( 2^{\lfloor \log_2 \frac{\sqrt{(m^2 + 6mn + n^2)} - (m + n)}{2} \rfloor}, \quad r_{\text{max}} \right)}.
    \label{eq:r}
\end{equation}
The constraint to ranks that are powers of \( 2 \) ensures compatibility with hardware acceleration requirements.
%
Additionally, we impose an upper bound \( r_{\text{max}} \) to ensure that \( C \) and \( R \) are significantly-low-rank matrices.

\begin{figure}[!t]
    \footnotesize
    \centering
    \begin{subfigure}{\linewidth}
        \centering
        \includegraphics[width=\linewidth]{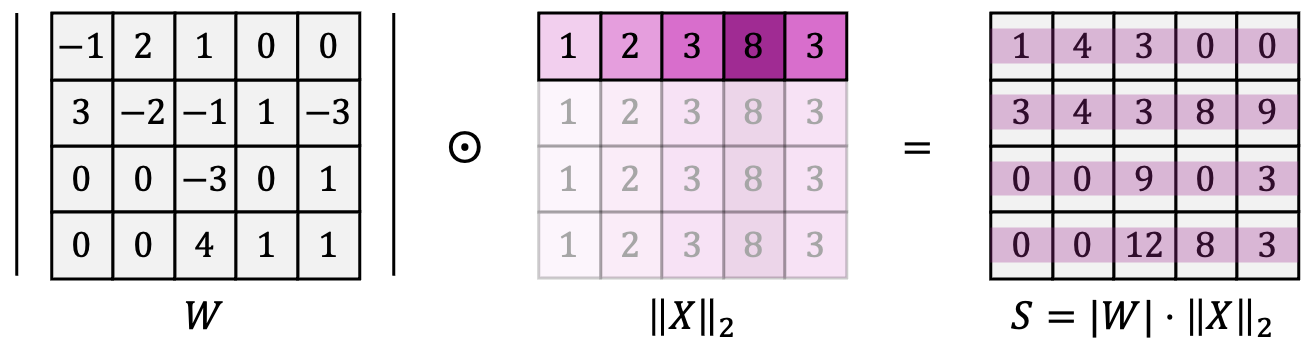}
        \caption{
            WANDA:
            The importance \(S\) is calculated as \( S = |W| \cdot \| X \|_2 \), given a weight matrix \(W\) and input feature activations \(X\).
        }
        \label{fig:process-1-subfig1}
    \end{subfigure}

    \vspace{8pt}

    \begin{subfigure}{\linewidth}
        \centering
        \includegraphics[width=\linewidth]{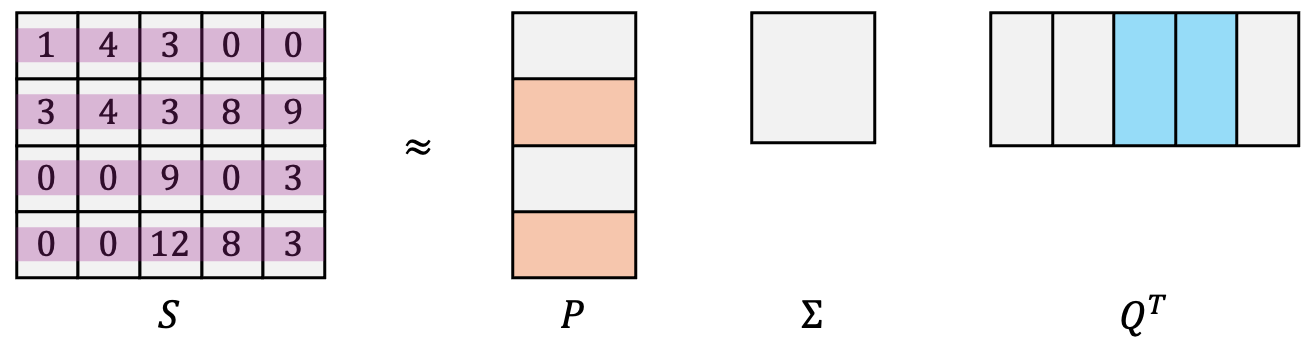}
        \caption{
            Using \(S\), the row and column indexes are selected via DEIM.
        }
        \label{fig:process-1-subfig2}
    \end{subfigure}

    \vspace{8pt}

    \begin{subfigure}{\linewidth}
        \centering
        \includegraphics[width=\linewidth]{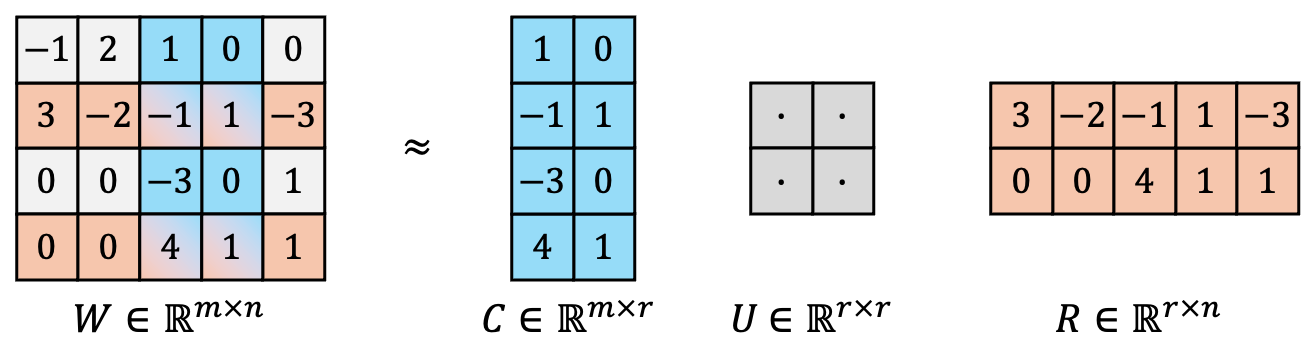}
        \caption{
            By selected indexes, \(C\) and \(R\) are extracted from the weight matrix \(W\), then \(U\) is computed as \(U = C^\dagger W R^\dagger\).
        }
        \label{fig:process-1-subfig3}
    \end{subfigure}
    \caption{
        Process of rank-\( r \) CUR decomposition in CURing.
        %
    }
    \label{fig:curing-process-1}
\end{figure}

\section{CURing} \label{section:CURing}

\begin{figure*}[!t]
    \footnotesize
    \centering
    \begin{subfigure}{0.20\textwidth}
        \centering
        \includegraphics[width=\linewidth]{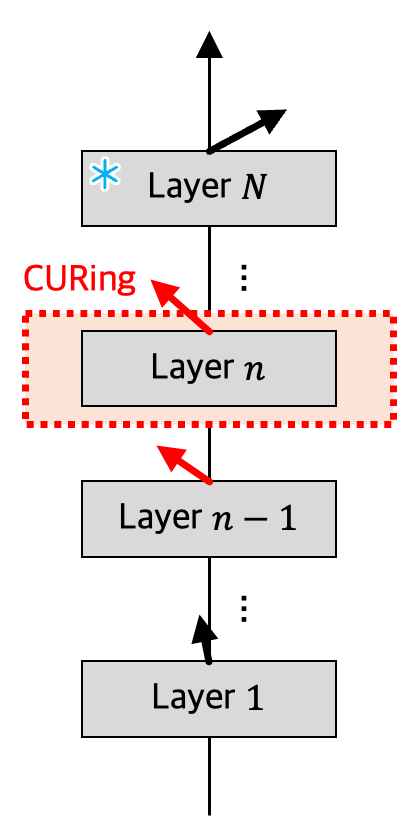}
        \caption{
            Layer Selection
        }
        \label{fig:process-2-subfig-1}
    \end{subfigure}
    \hspace{8pt}
    \rulesep
    \hspace{8pt}
    \begin{subfigure}{0.20\textwidth}
        \centering
        \includegraphics[width=\linewidth]{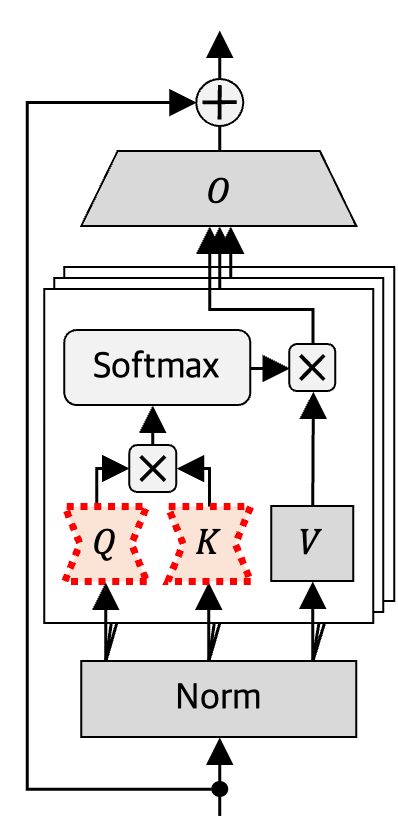}
        \caption{
            Decomposing MHA
        }
        \label{fig:process-2-subfig-2}
    \end{subfigure}
    \hspace{8pt}
    \rulesep
    \hspace{8pt}
    \begin{subfigure}{0.20\textwidth}
        \centering
        \includegraphics[width=\linewidth]{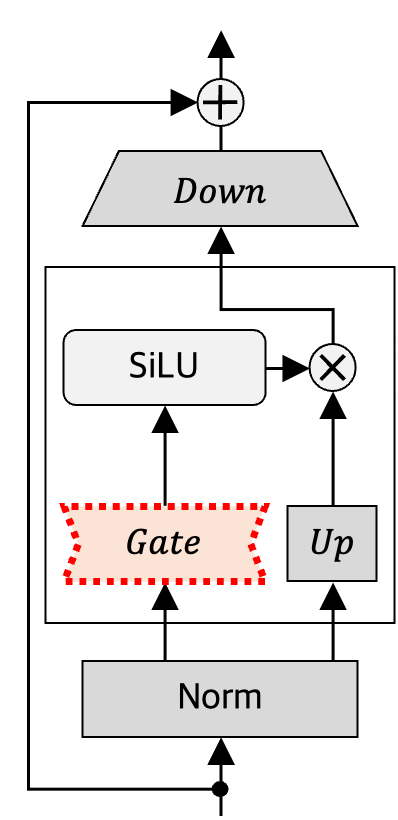}
        \caption{
            Decomposing FFN
        }
        \label{fig:process-2-subfig-3}
    \end{subfigure}
    \hspace{8pt}
    \rulesep
    \hspace{8pt}
    \begin{subfigure}{0.20\textwidth}
        \centering
        \includegraphics[width=\linewidth]{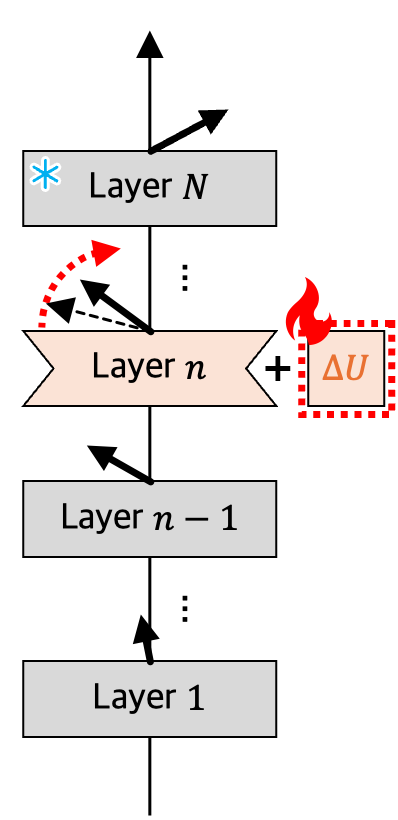}
        \caption{
            Layer-wise KD
        }
        \label{fig:process-2-subfig-4}
    \end{subfigure}
    \caption{
        CURing process illustrated based on the Llama3.1 architecture.
        (a) selecting target layers by angular distance, (b--c) decomposing their weights, and optionally (d) healing compression damage.
        The square multiplication symbol represents matrix multiplication, while the circular one denotes element-wise multiplication.
    }
    \label{fig:process-2}
\end{figure*}

Our proposed method, \textit{CURing}, compresses deep neural networks by reducing the rank of certain layers' weights using CUR matrix decomposition.
By identifying layers that contribute less to the model's performance, we replace their weights with low-rank approximations, significantly reducing model size without substantial loss in functionality.

We are focused on applying CURing to compress transformers~\cite{vaswani2017attention}, which are the foundation of most large-scale models.
Specifically, we focus on factorizing the Multi-Head Attention (MHA) and Feed-Forward Network (FFN) components of the transformer.
%
We apply compression specifically targeting the \textit{Query}, \textit{Key}, and \textit{Gate} weights in the Llama architecture (See the Figure~\ref{fig:process-2-subfig-2} and \ref{fig:process-2-subfig-3}).
%
We ignore biases for simplicity.
%
%
%
%
Appendix~\ref{appendix:weight-selection} briefly describes the effect of other combination of weights selection.
%

\subsection{Layer Selection} \label{section:layer-selection}

Several researches indicate that layers not playing significant roles can be removed in LLMs~\cite{gromov2024unreasonable, gardinazzi2024persistent}.
That means, we can replace them with low-rank representations, without loss of significant performance damage.
We focus on reducing the rank of layers exhibiting minimal changes—specifically, where the distance between their outputs is small.

To measure representation similarity, we compute the angular distance between the output representations of a layer and those of a subsequent layer.
The angular distance between two hidden states of layer \( n - 1 \) and \( n \), \( \mathbf{h}_{n-1} \) and \( \mathbf{h}_{n} \), is defined as:
\begin{equation}
    d(\mathbf{h}_{n-1}, \mathbf{h}_{n}) = \frac{1}{\pi} \arccos \left( \frac{\mathbf{h}_{n-1} \cdot \mathbf{h}_{n}}{\| \mathbf{h}_{n-1} \|_2 \| \mathbf{h}_{n} \|_2} \right),
    \label{eq:angular_distance}
    \nonumber
\end{equation}
where \( \cdot \) is the inner product over the hidden states of the last non-padded token of the sequence, and \( \| \cdot \|_2 \) denotes the \( \ell_2 \)-norm.
The hidden states are obtained and averaged over all calibration data.
In experiments, we use 128 Colossal Clean Crawled Corpus (C4) dataset~\cite{2020t5}.

Small distance implies that layers maintain similar information; thus, the later layer can be replaced with a low-rank approximation without significantly affecting performance.
In other words, for two similar layers, we perform CUR decomposition on the later layer, as shown in the Figure~\ref{fig:process-2-subfig-1}.
However, we retain the model's last layer, as it is essential for maintaining performance~\cite{gromov2024unreasonable}.

\subsection{CUR Decomposition on Weights} \label{subsection:cur-weights}

After selecting the layers, we compress the weights in each layer using CUR decomposition.
%
%
%
For each weight, to select rows and columns for CUR decomposition, we employ the \textit{WANDA}~\cite{sun2023simple} alongside the \textit{Discrete Empirical Interpolation Method} (DEIM)~\cite{sorensen2016deim}.

WANDA utilizes both weight magnitudes and activation information, advancing the selection criterion from a basic approach (considering only magnitudes) to a more sophisticated one (additionally considering changes).
%
%
%
%
As illustrated in Figure~\ref{fig:process-1-subfig1}, the information matrix $S$ is computed by multiplying the absolute values of the weights with the input activations.
%
This enriched information allows for the sensitive detection of fine weight influences, enabling effective pruning.
%
%
%
During calibration, we collect input activations concurrently as we compute per-layer angular distances using the same 128 C4 dataset.

In DEIM-CUR, for a given target rank \( r \),
the indices of the most important \( r \) rows and \( r \) columns are selected based on the Singular Value Decomposition (SVD) of informative matrix \( S \approx P \Sigma Q^T \).
This process is illustrated in Figure~\ref{fig:process-1-subfig2}.
%
%
%
%
%
%
Using the selected indexes, we extract \( C \) and \( R \) from the original matrix \( W \) (Figure~\ref{fig:process-1-subfig3}).
We then compute the core matrix \( U \) to approximate the original weight matrix \( W \).
The core matrix \( U \) is calculated using the pseudoinverses of \( C \) and \( R \), following the Equation~\ref{eq:U}.

Consider a simple fully connected (FC) network defined as:
\begin{equation}
    f_{W,W_2}(x) = \gamma(xW)W_2,
    \label{eq:fc}
\end{equation}
where \(\gamma\) is the activation function and biases are omitted for simplicity.
For convenience, we write \( f_W \) instead of \( f_{W,W_2} \).
We assume that the FC layer has sufficiently many hidden units.
Moreover, let any continuous \( f \in C(K) \) be defined on a compact set \( K \subset \mathbb{R}^m \) (thus \( x \in K \)), and let \(\gamma(\cdot)\) be \(L\)-Lipschitz continuous for some real constant \( L \ge 0 \):
\begin{equation}
    \| \gamma(a) - \gamma(b) \|_2 \le L \| a - b \|_2 \quad \text{for all } a,b.
    \label{eq:lipschitz}
\end{equation}
Under these conditions, and building upon prior findings on approximation errors~\cite{hajimolahoseini2021compressing,hornik1991approximation,eckart1936approximation}, the CUR-factorized network \( f_{CUR} \) satisfies the following error bound:
\begin{theorem}
    Let \( f_{CUR} \in C(K) \) be defined on a compact set \( K \subset \mathbb{R}^m \) with an \(L\)-Lipschitz activation \(\gamma\).
    Suppose a rank-\(r\) DEIM-CUR factorization (\( W \approx C U R \)) is applied to the fully connected layer.
    Then \( f_{CUR} \) approximates any continuous \( f \in C(K) \) within an error bound of \( (\epsilon + \delta) \):
    \[
        \| f - f_{CUR} \|_2^2 \le (\epsilon + \delta)^2,
    \]
    if the following inequality is satisfied:
    \[
        \sigma_{r+1} \le \frac{\delta}{L (\eta_p + \eta_q)} \left( \| W_2 \|_2 \| K \|_2 \right)^{-1},
    \]
    where \(\eta_p\) and \(\eta_q\) are finite error constants from Theorem~\ref{theorem:deim}, and \(\sigma_{r+1}\) is the \((r+1)\)-th singular value of the original matrix \( W \) (i.e., the first neglected singular value).
    \label{theorem:bound}
\end{theorem}
Here, \(\epsilon\) is the universal approximation error~\cite{hornik1991approximation} associated with the full-rank matrix \( W \), and \(\delta\) is the additional error introduced by the low-rank (\(r\)) CUR decomposition.
The proof of Theorem~\ref{theorem:bound} is provided in Appendix~\ref{appendix:proof}.

By applying CUR factorization to weights before activations
(i.e., \textit{Query}, \textit{Key}, and \textit{Gate} weights),
MHA and FFN layers can be viewed as FC-like structures, allowing the same approximation error bounds to hold.

\subsection{Decomposing Multi-Head Attentions}

In Transformer architectures, the Multi-Head Attention (MHA) mechanism plays a critical role in capturing contextual relationships within the input sequence~\cite{vaswani2017attention}.
Each MHA layer consists of multiple attention heads, each with its own set of query (\( Q \)), key (\( K \)), and value (\( V \)).
Given an input sequence \( X \in \mathbb{R}^{l \times d_{\text{model}}} \), where \( l \) is the sequence length and \( d_{\text{model}} \) is the model dimension, the weight matrices for the \( i \)-th head are defined as:
\[
W_i^{Q},\ W_i^{K},\ W_i^{V} \in \mathbb{R}^{d_{\text{model}} \times d_k},
\]
where \( d_k \) is the dimension of the queries and keys.
%
%
%
%
To simplify and align with the Llama architecture~\cite{dubey2024llama}, we set the hidden dimension of \( W^{V} \) to \( d_k \), the same as that of queries and keys.
Now, the queries, keys, and values are computed by projecting the input \( X \) using the weight matrices:
\[
Q_i = X W_i^{Q}, K_i = X W_i^{K}, V_i = X W_i^{V} \in \mathbb{R}^{l \times d_k}.
\]
For each attention head, the output is computed using the attention mechanism as follows:
\begin{equation}
\begin{split}
    \text{Head}_i(X)
    &= \text{Attention}(Q_i, K_i, V_i)\\
    &= \underbrace{\text{Softmax}\left( \frac{Q_i K_i^\top}{\sqrt{d_k}} \right)}_{P_{\text{Head}}} V_i.
    \nonumber
\end{split}
\end{equation}
The transformer heavily relies on the \( P_{\text{Head}} \in \mathbb{R}^{l \times l} \) part in attention to understand the context, by utilizing all tokens in the input sequence~\cite{wang2020linformer}.
Finally, the MHA output is obtained by concatenating the outputs of all heads and applying an output weight matrix \( W^{O} \in \mathbb{R}^{(h \cdot d_k) \times d_{\text{model}}} \):
\begin{equation}
\text{MHA}(X) = \text{Concat}\left( \text{Head}_1(X), \ldots, \text{Head}_h(X) \right) W^{O},
\nonumber
\end{equation}
where \( h \) is the number of heads.

To compress the MHA layers, we apply CUR decomposition specifically to the \( W^Q \) and \( W^K \) matrices.
Figure~\ref{fig:process-2-subfig-2} illustrates the decomposition of MHA.
%
We perform DEIM-CUR decomposition, taking their activations into consideration (WANDA):
\[
W_i^{Q} \approx C^Q_i U^Q_i R^Q_i, \quad W_i^{K} \approx C^K_i U^K_i R^K_i.
\]
Here, the rank-\(r\) factorized matrices \( C^Q_i, C^K_i \in \mathbb{R}^{d_{\text{model}} \times r} \) consist of selected columns from \( W_i^{Q} \) and \( W_i^{K} \), respectively.
Similarly, \( R^Q_i, R^K_i \in \mathbb{R}^{r \times d_k} \) consist of selected rows.
The core matrices \( U^Q_i, U^K_i \in \mathbb{R}^{r \times r} \) are computed using Equation~\ref{eq:U}.
%
%
After obtaining \( C^{\{Q, K\}}_i \), \( U^{\{Q, K\}}_i \), \( R^{\{Q, K\}}_i \), we compute the compressed queries and keys as:
\[
\widehat{Q_i} = X (C^Q_i U^Q_i R^Q_i), \quad \widehat{K_i} = X (C^K_i U^K_i R^K_i).
\]
Thus, \( Q_i \approx \widehat{Q_i} \) and \( K_i \approx \widehat{K_i} \).
The attention computation proceeds using \( \widehat{Q_i} \) and \( \widehat{K_i} \):
\begin{equation}
\begin{split}        
    \text{Head}_i(X)
    &= \text{Attention}(\widehat{Q_i}, \widehat{K_i}, V_i)\\
    &= \underbrace{\text{Softmax}\left( \frac{\widehat{Q_i} \widehat{K_i}^\top}{\sqrt{d_k}} \right)}_{\widehat{P_{\text{Head}}}} V_i.
    \nonumber
\end{split}
\end{equation}
Intuitively, \( \widehat{P_{\text{Head}}} \) represents a significant rank reduction in the matrix for context interpretation in transformers, based on the idea that layers causing minimal output changes do not require high context interpretation capacity.
%

\subsection{Decomposing Feed-Forward Networks}

%
We consider a Feed-Forward Network (FFN) to consist of a Gate, Up, and Down projections~\cite{shazeer2020glu,dauphin2017language}, with corresponding weights \( W^{\text{Gate}} \), \( W^{\text{Up}} \), and \( W^{\text{Down}} \).
Figure~\ref{fig:process-2-subfig-3} represents the structure of the FFN and presents a brief overview of its decomposition.

Given an input vector \( X \in \mathbb{R}^{l \times d_{\text{model}}} \), the FFN output is calculated as:
\begin{equation}
    \text{FFN}(X) = \left( \underbrace{\text{SiLU}(X W^{\text{Gate}})}_{P_{\text{FFN}}} \odot X W^{\text{Up}} \right) W^{\text{Down}},
    \nonumber
\end{equation}
where \( W^{\text{Gate}}, W^{\text{Up}} \in \mathbb{R}^{d_{\text{model}} \times d_{\text{inter}}} \) and \( W^{\text{Down}} \in \mathbb{R}^{d_{\text{inter}} \times d_{\text{model}}} \), with \( d_{\text{inter}} \) representing the intermediate dimension.
Llama employs the SiLU activation function~\cite{shazeer2020glu,ramachandran2017searching}.
The gate projection part, \(P_{\text{FFN}}\), effectively controls the flow of information.

To compress the FFN, we apply CUR decomposition to the weight \( W^{\text{Gate}} \). Specifically, we approximate this weight matrix as:
\[
W^{\text{Gate}} \approx C^{\text{Gate}} U^{\text{Gate}} R^{\text{Gate}},
\]
where \( C^{\text{Gate}} \in \mathbb{R}^{d_{\text{model}} \times r} \) contains selected columns from \( W^{\text{Gate}} \);
\( R^{\text{Gate}} \in \mathbb{R}^{r \times d_{\text{inter}}} \) consists of selected rows;
and \( U^{\text{Gate}} \in \mathbb{R}^{r \times r} \) is the core matrix calculated using Equation~\ref{eq:U}.
With this approximation of \(W^{\text{Gate}}\), \( \widehat{W^{\text{Gate}}} = C^{\text{Gate}} U^{\text{Gate}} R^{\text{Gate}} \), the FFN computation becomes:
\begin{equation}
    \text{FFN}(X) = \left( \underbrace{\text{SiLU}(X \widehat{W^{\text{Gate}}})}_{\widehat{P_{\text{FFN}}}} \odot X W^{\text{Up}} \right) W^{\text{Down}},
    \nonumber
\end{equation}
Applying CUR decomposition to \( W^{\text{Gate}} \) reduces the number of parameters in the FFN.
Since FFNs comprise roughly two-thirds of Transformer parameters~\cite{xia2022structured}, this also significantly reduces the overall model parameters.
%
%

%
%

\subsection{Layer-wise Knowledge Distillation} \label{subsection:healing}

Although retraining is not strictly required due to the inherent correction provided by CUR decomposition, additional training (healing) can be beneficial when the model undergoes substantial compression or when further performance improvement is desired.

%
In the healing process, we allow only \( U \) to be updated, while the matrices \( C \) and \( R \) remain fixed.
Further, the core matrix \( U \) is interpreted as \( U = U_0 + \Delta U \), where \( U_0 \) is initialized to the value of \( U \), and \( \Delta U \) starts as a zero matrix.
During healing, \( U_0 \) remains fixed while \(\Delta U\) is iteratively updated.
As illustrated in Figure~\ref{fig:adaptations}, this formulation allows healing to be intuitively viewed as a Parameter-Efficient Fine-Tuning (PEFT) method, where \( \Delta U \) corresponds to the trainable component.

However, the goal here is not task-specific adaptation as PEFT does, but to restore overall performance while mitigating catastrophic forgetting.
%
%
To achieve this, we use layer-wise Knowledge Distillation (KD), as illustrated in Figure~\ref{fig:process-2-subfig-4}.
We employ a layer-wise Mean-Squared-Error (MSE) loss between the teacher (original) and student (compressed) model outputs.
This approach aligns with techniques in prior work~\cite{sun2019patient, xia2022structured} and has proven effective in preserving the performance of compressed models~\cite{sreenivas2024llm, muralidharan2024compact}.
KD also acts as implicit regularization~\cite{tang2020understanding, saglietti2022solvable}, since the soft outputs of the teacher guide the student to prevent overfitting and constrain excessive parameter growth.
Therefore, catastrophic forgetting on previously learned information is mitigated, even if we use only one kind of corpus (e.g., C4) for healing.

%
A structural approach that fixes the \( C \) and \( R \) also mitigates forgetting.
Intuitively, only \( U \) is updated, and so the optimization is restricted to a subspace determined by \( C \) and \( R \).
Similar to MoRA’s \textit{comp}/\textit{decomp} modules~\cite{jiang2024mora}, \(C\) and \(R\) project parameters to and from a lower-rank space.
However, unlike MoRA, CURing’s fixed \(C\) and \(R\) impose additional constraints that regulate the update directions of \(U\).
This mitigates catastrophic forgetting, as observed in CURLoRA~\cite{fawi2024curlora}.

%
%
To analyze this more formally, let us revisit the single fully-connected network defined in Equation~\ref{eq:fc}, where the activation function \( \gamma(\cdot) \) is Lipschitz continuous with constant \( L \), as stated in Equation~\ref{eq:lipschitz}.
For an input batch \( X \) with \( b \) as the batch size, the MSE between the original output \( f_W(X) \) and its CUR approximation \( f_{CUR}(X) \) is given by:
\[
    \text{MSE} = \frac{1}{b}\|f_{W}(X) - f_{CUR}(X)\|_F^2.
\]
Meanwhile, we consider the Frobenius norm-based loss:  
\[
    \mathcal{L}(U) = \|W - C U R\|_F^2.
\]
Instead of considering the MSE directly, we use \( \mathcal{L}(U) \) in our analysis, as the MSE is upper-bounded by \( \mathcal{L}(U) \).
This allows us to analyze the network at the level of weight matrices, where minimizing \( \mathcal{L}(U) \) also optimizes the MSE.
\begin{theorem}
Let \( f_W \) and \( f_{CUR} \) represent the outputs of a fully connected network with weights \( W \) and their CUR factorized matrices \( C \), \( U \), and  \( R \), respectively.
Suppose the activation function \( \gamma(\cdot) \) is Lipschitz continuous with constant \( L \), and the input batch \( X \) (of size \( b \)) is sufficiently diverse and uniformly distributed.
Then, the MSE satisfies the following upper bound:
\begin{align}
    \text{MSE}(X) 
    &= \frac{1}{b}\|f_{W}(X) - f_{CUR}(X)\|_F^2 \nonumber \\
    &\leq \frac{1}{b} L^2 \|X\|_F^2 \|W_2\|_F^2 \mathcal{L}(U) \nonumber,
\end{align}
where \( \mathcal{L}(U) = \|W - C U R\|_F^2 \) is the Frobenius norm-based loss.
\label{theorem:loss}
\end{theorem}

Using \( \mathcal{L}(U) \), we can investigate the subspace restriction of \( U \) as follows:
\begin{theorem}
    Given \(W \approx CUR \in \mathbb{R}^{m \times n}\), let \( C \in \mathbb{R}^{m \times r} \) and \( R \in \mathbb{R}^{r \times n} \) be fixed, while \( U \in \mathbb{R}^{r \times r} \) is the only trainable matrix.
    Consider the loss function:
    \[
        \mathcal{L}(U) = \|W - C U R\|_F^2.
    \]
    The gradient of this loss with respect to \(U\), denoted as \( \nabla_{U} \mathcal{L}(U) \), always lies in the set:
    \[
        \nabla_{U} \mathcal{L}(U) \in \{C^\top M R^\top\},
    \]
    where \( M = CUR - W \in \mathbb{R}^{m \times n}\).
    \label{theorem:direction}
\end{theorem}
The proofs of Theorems~\ref{theorem:loss} and \ref{theorem:direction} can be found in Appendices~\ref{appendix:proof2} and \ref{appendix:proof3}, respectively.

%
We further consider the optimization problem:
\[
    U^\ast = \arg\min_{U} \mathcal{L}(U) = \arg\min_{U} \|W - C U R\|_F^2.
\]
The solution \(W^\ast = C U^\ast R\) is the best Frobenius norm approximation to \(W\), that is, \(\|W^\ast\|_F \approx \|W\|_F\).
As healing progresses to minimize \( \mathcal{L}(U) \), the compressed model’s representation \( CUR \) approaches \( W^\ast \).
%
Thus, scales becomes aligned with those of the original weights.
By constraining the update directions and scales, the healing process imposes structural regularization on the changes to \( U \) (i.e., restricting \(\Delta U\)).
Semantically, this enhances the context-interpreting performance of \( \widehat{P_{\text{Head}}} \) and \( \widehat{P_{\text{FFN}}} \) while mitigating forgetting.

Empirically, after KD, the Frobenius norm difference between \( W \) and \( CUR \) decreases, so the student's norm no longer overshoots the teacher's.
In experiments, we performed KD using the C4 dataset~\cite{2020t5}, excluding the data used for calibration in measuring layer-wise angular distances and accumulating WANDA input activations.
Remarkably, we observed that the model's performance across multiple tasks was quickly restored with only about 100 steps of fine-tuning.
This demonstrates the efficiency of our approach.
%
%

\definecolor{lightblue}{rgb}{0.3, 0.3, 1.0}
\definecolor{darkgreen}{rgb}{0.0, 0.7, 0.0}
\definecolor{lightred}{rgb}{1.0, 0.3, 0.3}

\begin{table*}[!t]
\footnotesize
\centering
\caption{
Performance comparison across various models and the number of compressed layers (\(r_\text{max} = 256\)).
}
\resizebox{\textwidth}{!}{%
\begin{tabular}{c*{16}{|r}}
\hline
\multirow{2}{*}{\textbf{Metrics}} & 
                \multicolumn{16}{c}{\textbf{Number of Compressed Layers}} \\ \cline{2-17}
                & \multicolumn{1}{c|}{\textbf{0}} & \multicolumn{1}{c|}{\textbf{2}}   & \multicolumn{1}{c|}{\textbf{4}}   & \multicolumn{1}{c|}{\textbf{6}}   & \multicolumn{1}{c|}{\textbf{8}}   & \multicolumn{1}{c|}{\textbf{10}}  & \multicolumn{1}{c|}{\textbf{12}}  & \multicolumn{1}{c|}{\textbf{14}}  & \multicolumn{1}{c|}{\textbf{16}}  & \multicolumn{1}{c|}{\textbf{18}}  & \multicolumn{1}{c|}{\textbf{20}}  & \multicolumn{1}{c|}{\textbf{22}}  & \multicolumn{1}{c|}{\textbf{24}}  & \multicolumn{1}{c|}{\textbf{26}}  & \multicolumn{1}{c|}{\textbf{28}}  & \multicolumn{1}{c}{\textbf{30}}  \\ \hline \hline
\rowcolor{lightblue!30}
\multicolumn{17}{c}{\textbf{Llama3.1-8B}} \\ \hline
\textbf{time (s)} &
                - & 33.62 & 59.14 & 85.94 & 104.95 & 129.45 & 154.75 & 177.59 & 204.40 & 228.16 & 249.32 & 280.85 & 300.54 & 325.90 & 358.14 & 376.29 \\ 
\textbf{params} &
                8.03B & 7.89B & 7.74B & 7.60B & 7.46B & 7.32B & 7.17B & 7.03B & 6.89B & 6.75B & 6.60B & 6.46B & 6.32B & 6.17B & 6.03B & 5.89B \\
\textbf{GiB} &
                {29.92} & $\blacktriangledown$0.53 & $\blacktriangledown$1.06 & $\blacktriangledown$1.60 & $\blacktriangledown$2.13 & $\blacktriangledown$2.66 & $\blacktriangledown$3.19 & $\blacktriangledown$3.72 & $\blacktriangledown$4.25 & $\blacktriangledown$4.79 & $\blacktriangledown$5.32 & $\blacktriangledown$5.85 & $\blacktriangledown$6.38 & $\blacktriangledown$6.91 & $\blacktriangledown$7.44 & $\blacktriangledown$7.98 \\ \hline \hline
%
\rowcolor{darkgreen!30}
\multicolumn{17}{c}{\textbf{Llama2-7B}} \\ \hline
\textbf{time (s)} &
                - & 36.35 & 57.96 & 82.90 & 109.88 & 135.70 & 160.50 & 185.12 & 208.97 & 235.51 & 264.06 & 288.32 & 313.38 & 344.71 & 368.67 & 395.53 \\ 
\textbf{params} &
                6.74B & 6.60B & 6.46B & 6.32B & 6.18B & 6.03B & 5.89B & 5.75B & 5.61B & 5.47B & 5.33B & 5.19B & 5.05B & 4.91B & 4.77B & 4.63B \\
\textbf{GiB} &
                {25.10} & $\blacktriangledown$0.52 & $\blacktriangledown$1.05 & $\blacktriangledown$1.57 & $\blacktriangledown$2.10 & $\blacktriangledown$2.62 & $\blacktriangledown$3.15 & $\blacktriangledown$3.67 & $\blacktriangledown$4.20 & $\blacktriangledown$4.72 & $\blacktriangledown$5.24 & $\blacktriangledown$5.77 & $\blacktriangledown$6.29 & $\blacktriangledown$6.82 & $\blacktriangledown$7.34 & $\blacktriangledown$7.87 \\ \hline \hline
%
\rowcolor{lightred!30}
\multicolumn{17}{c}{\textbf{Mistral-7B}} \\ \hline
\textbf{time (s)} &
                - & 34.18 & 59.38 & 82.60 & 106.96 & 132.96 & 153.03 & 177.83 & 201.50 & 233.79 & 249.82 & 279.67 & 300.20 & 329.95 & 349.88 & 383.58 \\
\textbf{params} &
                7.24B & 7.10B & 6.96B & 6.81B & 6.67B & 6.53B & 6.39B & 6.24B & 6.10B & 5.96B & 5.81B & 5.67B & 5.53B & 5.39B & 5.24B & 5.10B \\
\textbf{GiB} &
                {26.98} & $\blacktriangledown$0.53 & $\blacktriangledown$1.06 & $\blacktriangledown$1.60 & $\blacktriangledown$2.13 & $\blacktriangledown$2.66 & $\blacktriangledown$3.19 & $\blacktriangledown$3.72 & $\blacktriangledown$4.25 & $\blacktriangledown$4.79 & $\blacktriangledown$5.32 & $\blacktriangledown$5.85 & $\blacktriangledown$6.38 & $\blacktriangledown$6.91 & $\blacktriangledown$7.44 & $\blacktriangledown$7.98 \\ \hline \hline
%
\rowcolor{cyan!30}
\multicolumn{17}{c}{\textbf{Orca2-7B}} \\ \hline
\textbf{time (s)} &
                - & 34.38 & 59.76 & 86.75 & 114.61 & 139.82 & 166.95 & 195.10 & 219.17 & 248.80 & 260.56 & 301.34 & 322.24 & 354.96 & 384.29 & 408.58 \\ 
\textbf{params} &
                6.74B & 6.60B & 6.46B & 6.32B & 6.18B & 6.03B & 5.89B & 5.75B & 5.61B & 5.47B & 5.33B & 5.19B & 5.05B & 4.91B & 4.77B & 4.63B \\
\textbf{GiB} &
                {25.10} & $\blacktriangledown$0.52 & $\blacktriangledown$1.05 & $\blacktriangledown$1.57 & $\blacktriangledown$2.10 & $\blacktriangledown$2.62 & $\blacktriangledown$3.15 & $\blacktriangledown$3.67 & $\blacktriangledown$4.20 & $\blacktriangledown$4.72 & $\blacktriangledown$5.24 & $\blacktriangledown$5.77 & $\blacktriangledown$6.29 & $\blacktriangledown$6.82 & $\blacktriangledown$7.34 & $\blacktriangledown$7.87 \\ \hline
\end{tabular}%
}
\label{tab:layers}
\end{table*}
\input{texts/5_2_2_comp_graph}

\section{Experiments} \label{section:experiments}

%
We evaluate \textit{CURing} across multiple datasets and settings to assess its compression efficiency, performance retention, and healing capabilities.
Calibration (calculating WANDA and angular distances) and healing data are drawn from the C4 training set, with no overlap between the two.
The use of C4 provides superior performance in calibration compared to other corpora~\cite{flynn2024stat,sun2023simple}.
By default, we calibrate on 128 examples.
We evaluate models on the C4 validation subset, WikiText2, BoolQ (0-shot), and MMLU (5-shot).
For MMLU, we use 32 samples per category from 57 categories.
The context length is capped at 128;
%
%
the detailed hyperparameters appear in Appendix~\ref{appendix:hyperparameters}.
All experiments are conducted on a single NVIDIA H100 80GB GPU.

\subsection{Compression Performance}

We apply CURing to multiple models—Llama3.1-8B, Llama2-7B, Mistral-7B, and Orca2-7B—to investigate compression overhead, size reduction, and performance impact.
%
%
%
%

%
%
Table~\ref{tab:layers} presents the time required for CURing as we vary the number of compressed layers (\(1\) to \(30\)), excluding the first and last layers as discussed in Section~\ref{section:layer-selection}.
The max rank is fixed at \(256\) (other ranks are discussed in Appendix~\ref{appendix:rank}).
%
%
%
%
%
%
%
Our CURing is significantly faster than other compressing methods, such as SliceGPT~\cite{ashkboos2024slicegpt}, which requires about 44 minutes to prune Llama2-7B to a size of 6.11B on a single H100.
Similarly, STAT~\cite{flynn2024stat} and LLM Surgeon~\cite{van2023llm} require anywhere from tens of minutes to several hours.
Most of these overheads arise from structural factors, such as structural pruning or handling residual connections
In comparison, CURing achieves a similar compression level—for instance, compressing 10 layers—in just about 2 minutes.
By leveraging the efficient CUR decomposition and avoiding complex structural considerations, CURing achieves remarkable speed.
%
%
%
%

As also detailed in Table~\ref{tab:layers},
%
%
obviously, increasing the number of compressed layers proportionally reduces the model size.
However, greater compression results in a more negative impact on performance.
%
%
%
%
%
%
%
%
%
%
%
%
%
As shown in Figure~\ref{fig:performance}, we report perplexity on C4 and WikiText2, along with zero-shot or few-shot accuracies on BoolQ and MMLU.
For Llama3.1-8B, the uncompressed baseline (\(x=0\)) yields perplexity values of \(23.79\) on C4 and \(566.21\) on WikiText2, with BoolQ at \(82.11\%\) and MMLU at \(67.32\%\).
After compressing 10 layers at \(r_{\text{max}}=256\) and without healing, perplexity becomes \(77.33\) on C4 and \(705.40\) on WikiText2, while BoolQ and MMLU accuracies are \(75.78\%\) and \(64.91\%\).
Similar tendencies emerge for Mistral-7B and Orca2-7B.
Although some performance reduction occurs, the models still markedly outperform random baselines.
Empirically, compressing roughly 9--11 layers strikes a good balance between size savings and accuracy, while compressing even more than half of the layers still exceeds random baselines.


\input{texts/5_3_2_heal_graph}
\renewcommand{\thesubfigure}{Figure \thefigure}
\captionsetup[subfigure]{labelformat=simple, labelsep=colon, font=normalsize}

\definecolor{darkgreen}{rgb}{0.0, 0.7, 0.0}
\definecolor{lightred}{rgb}{1.0, 0.3, 0.3}

\begin{figure*}[!t]
\centering
\footnotesize

\begin{center}
    \pgfplotslegendfromname{forgetnamedlegend}
\end{center}

\resizebox{\textwidth}{!}{ %
\begin{tikzpicture}
\begin{groupplot}[
    group style={
        group size=5 by 1,
        horizontal sep=1.0cm,
    },
    width=5.1cm,
    height=4.2cm,
    xmin=-0.15, xmax=4.15,
    ymin=0.55, ymax=0.95,
    xtick={0, 1.0, 2.0, 3.0, 4.0},
    xticklabel={\pgfmathparse{\tick}\pgfmathprintnumber{\pgfmathresult}k},
    xtick align=outside,
    xtick pos=bottom,
    ylabel=\empty,
    ytick={0.6, 0.7, 0.8, 0.9},
    ytick align=outside,
    ytick pos=left,
]

%
\nextgroupplot[
    title={MRPC Train Loss},title style={yshift=-0.2cm},
    ylabel=\empty,
    ymin=0.0, ymax=6.0,
    ytick={0, 1, 2, 3, 4, 5, 6},
    legend style={
        column sep=0.15cm,
        legend columns=4,
    },
    legend cell align={center},
    legend to name=forgetnamedlegend
]
\addplot[blue, mark=*] coordinates {
(0.0, 5.699847) (0.25, 2.558288) (0.5, 2.375626) (0.75, 2.311156) (1.0, 2.029808) (1.25, 2.049947) (1.5, 2.067731) (1.75, 2.242374) (2.0, 1.814613) (2.25, 1.739576) (2.5, 1.727538) (2.75, 1.765248) (3.0, 1.522884) (3.25, 1.569011) (3.5, 1.55832) (3.75, 1.544716) (4.0, 1.494077)
};
\addlegendentry{CURing}
\addplot[darkgreen, mark=triangle*] coordinates {
(0.0, 5.927737) (0.25, 2.419984) (0.5, 2.108674) (0.75, 1.97084) (1.0, 1.320313) (1.25, 1.52555) (1.5, 1.178342) (1.75, 1.081601) (2.0, 0.642411) (2.25, 0.419904) (2.5, 0.395217) (2.75, 0.329416) (3.0, 0.302592) (3.25, 0.378286) (3.5, 0.311046) (3.75, 0.237648) (4.0, 0.282184)
};
\addlegendentry{LoRA}
\addplot[lightred, mark=x] coordinates {
(0.0, 5.700062) (0.25, 2.459898) (0.5, 1.857046) (0.75, 1.870492) (1.0, 1.638175) (1.25, 1.071574) (1.5, 1.128072) (1.75, 0.933318) (2.0, 0.79568) (2.25, 0.639952) (2.5, 0.581996) (2.75, 0.437307) (3.0, 0.491329) (3.25, 0.414978) (3.5, 0.334611) (3.75, 0.339404) (4.0, 0.365661)
};
\addlegendentry{MoRA}
\addplot[cyan, mark=diamond*] coordinates {
(0.0, 5.678503) (0.25, 2.815717) (0.5, 2.827884) (0.75, 2.850643) (1.0, 2.168583) (1.25, 2.578858) (1.5, 2.38658) (1.75, 2.451201) (2.0, 1.958095) (2.25, 2.351893) (2.5, 2.48118) (2.75, 2.222063) (3.0, 2.966647) (3.25, 2.709349) (3.5, 2.32127) (3.75, 2.301009) (4.0, 2.404836)
};
\addlegendentry{CURLoRA}
\addplot[gray!30] coordinates {(0,0.0) (0,7.0)};
\addplot[gray!30] coordinates {(1.0,0.0) (1.0,7.0)};
\addplot[gray!30] coordinates {(2.0,0.0) (2.0,7.0)};
\addplot[gray!30] coordinates {(3.0,0.0) (3.0,7.0)};
\addplot[gray!30] coordinates {(4.0,0.0) (4.0,7.0)};

%
\nextgroupplot[
    title={WikiText2 PPL},title style={yshift=-0.2cm},
    ylabel=\empty,
    ymode=log,
    ymin=40.0, ymax=2000.0,
    ytick={10, 100, 1000, 10000},
]
\addplot[gray!30] coordinates {(0.0,40.0) (0.0,2000.0)};
\addplot[gray!30] coordinates {(1.0,40.0) (1.0,2000.0)};
\addplot[gray!30] coordinates {(2.0,40.0) (2.0,2000.0)};
\addplot[gray!30] coordinates {(3.0,40.0) (3.0,2000.0)};
\addplot[gray!30] coordinates {(4.0,40.0) (4.0,2000.0)};
\addplot[blue, mark=*] coordinates {
(0.0, 737.236) (0.25, 98.36525) (0.5, 112.5655) (0.75, 150.51225) (1.0, 176.292) (1.25, 226.40075) (1.5, 270.16) (1.75, 305.6955) (2.0, 362.106) (2.25, 392.30025) (2.5, 432.6625) (2.75, 443.8925) (3.0, 441.381) (3.25, 538.274) (3.5, 593.336) (3.75, 646.23425) (4.0, 675.608)
};
\addplot[darkgreen, mark=triangle*] coordinates {
(0.0, 737.235) (0.25, 51.36175) (0.5, 84.3655) (0.75, 152.395) (1.0, 218.633) (1.25, 339.00225) (1.5, 431.1885) (1.75, 566.403) (2.0, 668.251) (2.25, 823.616) (2.5, 1019.8195) (2.75, 1011.89475) (3.0, 1044.712) (3.25, 1295.2035) (3.5, 1284.9615) (3.75, 1357.13025) (4.0, 1407.289)
};
\addplot[lightred, mark=x] coordinates {
(0.0, 737.235) (0.25, 124.435) (0.5, 229.679) (0.75, 297.85725) (1.0, 383.849) (1.25, 468.914) (1.5, 489.675) (1.75, 588.2875) (2.0, 688.693) (2.25, 847.175) (2.5, 925.091) (2.75, 967.4315) (3.0, 820.291) (3.25, 834.38075) (3.5, 1013.4195) (3.75, 1113.6835) (4.0, 1022.905)
};
\addplot[cyan, mark=diamond*] coordinates {
(0.0, 737.235) (0.25, 148.59925) (0.5, 145.821) (0.75, 150.46925) (1.0, 154.651) (1.25, 152.46225) (1.5, 153.105) (1.75, 143.8875) (2.0, 138.507) (2.25, 137.01475) (2.5, 136.664) (2.75, 133.38825) (3.0, 131.088) (3.25, 128.70175) (3.5, 125.798) (3.75, 129.1485) (4.0, 128.036)
};

\nextgroupplot[
    axis lines=none,
    ticks=none,
    width=1.6cm
]

%
%

%
\nextgroupplot[
    title={UUID Train Loss},title style={yshift=-0.2cm},
    xmin=-0.5, xmax=12.5,
    xtick={0, 2, 4, 6, 8, 10, 12},
    xticklabel={\pgfmathparse{\tick}\pgfmathprintnumber{\pgfmathresult}k},
    xtick align=outside,
    xtick pos=bottom,
    ylabel=\empty,
    ymode=log,
    ymin=0.04, ymax=20.0,
    ytick={0.01, 0.1, 1, 10, 100},
    legend style={
        column sep=0.15cm,
        legend columns=4,
    },
    legend cell align={center},
    legend to name=uuidnamedlegend
]
\addplot[blue, mark=*] coordinates {
    (0, 10.639) (1, 1.394) (2, 1.331) (3, 1.286) (4, 1.137)
    (5, 0.814) (6, 0.587) (7, 0.347) (8, 0.211) (9, 0.153)
    (10, 0.136) (11, 0.089) (12, 0.080)
};
\addlegendentry{CURing}

\addplot[darkgreen, mark=triangle*] coordinates {
    (0, 10.830) (1, 1.189) (2, 0.121) (3, 0.085) (4, 0.065)
    (5, 0.060) (6, 0.067) (7, 0.060) (8, 0.056) (9, 0.059)
    (10, 0.058) (11, 0.062) (12, 0.058)
};
\addlegendentry{LoRA}

\addplot[lightred, mark=x] coordinates {
    (0, 10.811) (1, 1.356) (2, 0.835) (3, 0.222) (4, 0.104)
    (5, 0.084) (6, 0.083) (7, 0.073) (8, 0.070) (9, 0.072)
    (10, 0.078) (11, 0.073) (12, 0.071)
};
\addlegendentry{MoRA}

\addplot[cyan, mark=diamond*] coordinates {
    (0, 10.535593032836900)
    (1, 1.467537522315980)
    (2, 1.4238619804382300)
    (3, 1.4194234609603900)
    (4, 1.412614107131960)
    (5, 1.4126391410827600)
    (6, 1.3989661931991600)
    (7, 1.3983068466186500)
    (8, 1.390293836593630)
    (9, 1.3830151557922400)
    (10, 1.3810361623764000)
    (11, 1.3711705207824700)
    (12, 1.3808962106704700)
};
\addlegendentry{CURLoRA}
\addplot[gray!30] coordinates {(0,0.01) (0,20)};
\addplot[gray!30] coordinates {(2,0.01) (2,20)};
\addplot[gray!30] coordinates {(4,0.01) (4,20)};
\addplot[gray!30] coordinates {(6,0.01) (6,20)};
\addplot[gray!30] coordinates {(8,0.01) (8,20)};
\addplot[gray!30] coordinates {(10,0.01) (10,20)};
\addplot[gray!30] coordinates {(12,0.01) (12,20)};

%
\nextgroupplot[
    title={UUID Accuracy},title style={yshift=-0.265cm},
    xmin=-0.5, xmax=12.5,
    xtick={0, 2, 4, 6, 8, 10, 12},
    xticklabel={\pgfmathparse{\tick}\pgfmathprintnumber{\pgfmathresult}k},
    xtick align=outside,
    xtick pos=bottom,
    ylabel=\empty,
    ymin=0.1, ymax=0.8,
    ytick={0.1, 0.2, 0.3, 0.4, 0.5, 0.6, 0.7, 0.8},
]
\addplot[blue, mark=*] coordinates {
    (0.0, 0.137) (1.0, 0.195) (2.0, 0.199) (3.0, 0.202) (4.0, 0.204) (5.0, 0.214) (6.0, 0.275) (7.0, 0.381) (8.0, 0.537) (9.0, 0.671) (10.0, 0.703) (11.0, 0.704) (12.0, 0.700)
};
\addplot[darkgreen, mark=triangle*] coordinates {
    (0.0, 0.137) (1.0, 0.2) (2.0, 0.648) (3.0, 0.7) (4.0, 0.701) (5.0, 0.705) (6.0, 0.7) (7.0, 0.708) (8.0, 0.643) (9.0, 0.708) (10.0, 0.707) (11.0, 0.709) (12.0, 0.709)
};
\addplot[lightred, mark=x] coordinates {
    (0.0, 0.137) (1.0, 0.196) (2.0, 0.226) (3.0, 0.639) (4.0, 0.767) (5.0, 0.767) (6.0, 0.767) (7.0, 0.765) (8.0, 0.77) (9.0, 0.764) (10.0, 0.768) (11.0, 0.769) (12.0, 0.762)
};
\addplot[cyan, mark=diamond*] coordinates {
    (0, 0.129584)
    (1, 0.185533)
    (2, 0.187971)
    (3, 0.192579)
    (4, 0.195597)
    (5, 0.194707)
    (6, 0.19574)
    (7, 0.197554)
    (8, 0.197835)
    (9, 0.198647)
    (10, 0.196316)
    (11, 0.194383)
    (12, 0.195584)
};
\addplot[gray!30] coordinates {(0,0.0) (0,1.0)};
\addplot[gray!30] coordinates {(2,0.0) (2,1.0)};
\addplot[gray!30] coordinates {(4,0.0) (4,1.0)};
\addplot[gray!30] coordinates {(6,0.0) (6,1.0)};
\addplot[gray!30] coordinates {(8,0.0) (8,1.0)};
\addplot[gray!30] coordinates {(10,0.0) (10,1.0)};
\addplot[gray!30] coordinates {(12,0.0) (12,1.0)};

\end{groupplot}
\end{tikzpicture} %
}


\begin{subfigure}[b]{0.495\textwidth}
    \caption{
        Comparisons of CURing and PEFT methods while training on MRPC.
        Perplexity is measured on WikiText2.
    }
    \label{fig:forget}
\end{subfigure}
\hfill
\stepcounter{figure}
\begin{subfigure}[b]{0.495\textwidth}
    \caption{
        Training loss and character-level accuracy for 1,024 new UUID-to-UUID mapping pairs.
    }
    \label{fig:uuid}
\end{subfigure}

\end{figure*}

\subsection{Healing}

To further enhance performance after compression, we conduct layer-wise Knowledge Distillation (KD) on C4 using Llama3.1-8B as the teacher and its compressed version as the student.
We run 2,000 steps (32,000 samples) of KD, setting \(\alpha=0.1\) so that the distillation loss is weighted by \((1-\alpha) = 0.9\).
As seen in Figure~\ref{fig:performance}, performance damage from compression is effectively recovered during healing, particularly for perplexities on C4 and WikiText2.
Notably, perplexity even improves beyond the original levels after healing: in the 10-layers case, C4 perplexity becomes to \(17.56\) with healing (far down from \(77.33\) without healing), even below the original Llama3.1-8B baseline of \(23.79\).
Despite retraining the CURing model on C4 alone, the performance on WikiText2 and other tasks also improves.
The WikiText2 perplexity drops to \(97.95\) with healing, compared to \(705.40\) without healing and \(566.21\) for the original model.

Figure~\ref{fig:healing} illustrates the healing process for a 10-layer-compressed model.
The performance rebounds quickly—often within the first 100 steps, demonstrating the efficiency of our approach in restoring.
We further compare our healing approach, which updates \(\Delta U\) in CURing, with two popular adaptation methods, LoRA~\cite{hu2021lora} and MoRA~\cite{jiang2024mora}.
We ensure that all methods have an equal number of trainable parameters.
%
%
Overall, all methods effectively restore performance.
However, on WikiText2, the CURing-based update achieves lower perplexity than LoRA.
We hypothesize this improvement is due to the relatively high-rank updates possible with CURing and MoRA, in contrast to LoRA’s inherently lower-rank structure.
On the other hand, the CURing update lags behind MoRA, possibly because updating \(\Delta U\) is restricted to the subspace defined by \( C\) and \( R\), as discussed in Section~\ref{subsection:healing}.
%
%
%


\section{Discussion}

%
\subsection{Interpretability}


Various methods have been proposed to interpret the behaviors of neural networks.
For instance, recent work~\cite{bricken2023monosemanticity} applies linear combinations of neuron activations (features) to isolate recurring activation patterns across diverse contexts.
Another approach~\cite{templeton2024scaling} quantifies the distance between features by identifying which neurons are appeared in activation patterns.
By retaining the principal components of the original weight matrices, CURing enables the reuse of existing interpretations about how the network processes information.
We observe that the activation levels in the selected columns of the compressed model (\( C \)) closely align with those in the original model, indicating that essential semantic information is preserved.
Appendix~\ref{appendix:activations} presents further details of these observations.



%
%
\subsection{Role of \( \Delta U \) in PEFT}

In CURing, the matrix \(\Delta U\) plays a role similar to Parameter-Efficient Fine-Tuning (PEFT) approaches such as LoRA~\cite{hu2021lora} and MoRA~\cite{jiang2024mora}, but with a distinct objective.
Whereas LoRA and MoRA focus on rapid adaptation to new tasks (often at a higher risk of forgetting), CURing’s primary goal is to restore and maintain the original model’s performance rather than accommodate new tasks.
%
%
%
%
To evaluate forgetting, we fine-tuned the model on MRPC for 4,000 steps while periodically evaluating its WikiText2 performance.
We compared CURing with LoRA and MoRA under the same learnable-parameter budget.
We also examined CURLoRA~\cite{fawi2024curlora}, a PEFT method specifically designed to address catastrophic forgetting.
As shown in Figure~\hyperref[fig:forget]{6}, LoRA and MoRA adapted to MRPC more quickly but showed larger increases in WikiText2 perplexity, indicating stronger forgetting of previously learned language modeling.
CURLoRA remained highly stable on WikiText2 but was almost unable to learn MRPC. 
These observations highlight a trade-off:
LoRA and MoRA achieve faster task adaptation but risk overwriting prior knowledge, while CURLoRA largely preserves the original model’s representations at the expense of new-task learning.
For healing, both learning capacity and memory retention are crucial.
CURing, by design, functions as a slower but more stable learner between them.
The subspace constraint provides a controlled path toward recovering the original performance without excessive forgetting.
%

%
%
Although the healing step in CURing is not primarily designed for learning entirely new tasks, we conducted a more detailed investigation on it.
By comparing CURing with other parameter-efficient fine-tuning (PEFT) methods, we aimed to gain a deeper understanding of its behaviors and characteristics.
Similar to the approach in MoRA, we created a random UUID-to-UUID mapping task of 1,024 pairs, providing data the model had never seen before.
As depicted in Figure~\hyperref[fig:uuid]{7}, CURing converged more slowly than LoRA and MoRA, ultimately matching LoRA’s performance with additional steps.
MoRA, benefiting from a higher-rank space, outperformed LoRA.
CURing also has a high-rank capacity but is restricted to the subspace defined by \( C \) and \( R \), preventing it from reaching MoRA’s final accuracy.
Still, it can at least match LoRA’s accuracy level, albeit more slowly.
Consequently, while CURing can learn new content, it is not as rapid for new-domain adaptation as MoRA or LoRA.
If faster adaptation to new tasks is the priority, LoRA or MoRA remain more suitable.
Conversely, CURing’s subspace restriction makes it an attractive option for scenarios like healing, where preserving previously acquired knowledge is crucial.
%

\section{Conclusion}

In this work, we presented \textit{CURing}, a model compression technique that leverages CUR matrix decomposition to effectively reduce neural network sizes while preserving performance, structural integrity, and interpretability.
Unlike traditional pruning methods that often require retraining, CURing achieves compression without necessitating additional training steps.
%
%
%
Our experimental results demonstrated that CURing substantially compresses models with minimal performance degradation across various tasks.
Analysis of activation patterns revealed that the internal representations of the compressed models closely align with those of the original models, maintaining interpretability.

Future research directions include exploring advanced decomposition techniques to further enhance efficiency and compactness.
For instance, methods like Compact Matrix Decomposition (CMD) or other approaches for sparse matrices proposed in prior research~\cite{sun2007less, ekenta2022spectrum} could yield more efficient low-rank factorization.

\section*{Acknowledgements}
The work is supported by \href{https://cplabs.io/en}{CPLABS, Inc}.


\bibliography{main}
\bibliographystyle{icml2025}


\newpage
\appendix
\onecolumn

\section{Proofs}


\subsection{CURing Approximation Error Bound} \label{appendix:proof}

We will show that the error of the DEIM-CUR factorized network is bounded as stated in Theorem~\ref{theorem:bound}.

\begin{proof}
    Consider \(\| f - f_{CUR} \|_2^2\), where \(\|\cdot\|_2\) denotes the \(\ell_2\)-norm over the domain \(K\):
    \[
        \| f - f_{CUR} \|_2^2 = \int_{K} \left(f(x) - f_{CUR}(x)\right)^2 d\mu.
    \]
    Decompose the integrand as:
    \[
        f(x) - f_{CUR}(x) = \left(f(x) - f_W(x)\right) + \left(f_W(x) - f_{CUR}(x)\right).
    \]
    Thus:
    \begin{subequations}
    \begin{align}
        \| f - f_{CUR} \|_2^2 &= \int_K \left(f(x)-f_W(x)\right)^2 d\mu \label{eq:proof-1}\\
        &\quad + \int_K \left(f_W(x)-f_{CUR}(x)\right)^2 d\mu \label{eq:proof-2}\\
        &\quad + 2 \int_K \left(f(x)-f_W(x)\right)\left(f_W(x)-f_{CUR}(x)\right)d\mu. \label{eq:proof-3}
    \end{align}
    \end{subequations}


    From the universal approximation theorem~\cite{hornik1991approximation}, since \( f_W \) is sufficiently expressive, we have \(\| f - f_W \|_2^2 \le \epsilon^2\). Therefore, the integral in \eqref{eq:proof-1} is bounded by \(\epsilon^2\).

    
    Next, consider the term in \eqref{eq:proof-2}. Using Equation~\ref{eq:fc}, we obtain:
    \[
        f_{W}(x) - f_{CUR}(x) = \gamma(xW)W_2 - \gamma(xCUR)W_2.
    \]
    By the Lipschitz continuity of \(\gamma(\cdot)\) (Equation~\ref{eq:lipschitz}), we have:
    \[
        \|\gamma(xW)-\gamma(xCUR)\|_2 \le L \|x(W - CUR)\|_2.
    \]
    Thus:
    \[
        \| f_{W}(x)-f_{CUR}(x) \|_2^2 = \|\gamma(xW)-\gamma(xCUR)\|_2^2 \|W_2\|_2^2 \le L^2 \|x\|_2^2 \|W_2\|_2^2 \|W - CUR\|_2^2.
    \]
    Since \(x \in K\) and \(K\) is compact, \(\|x\|_2 \le \|K\|_2\).
    Also, from Theorem~\ref{theorem:deim}, \(\|W-CUR\|_2 \le (\eta_p+\eta_q)\sigma_{r+1}\).
    Therefore:
    \[
        \int_{K} \left(f_{W}(x) - f_{CUR}(x)\right)^2 d\mu \le L^2 \|W_2\|_2^2 \|K\|_2^2 \left((\eta_p + \eta_q)\sigma_{r+1}\right)^2.
    \]


    For the cross-term \eqref{eq:proof-3}, apply the \textit{Cauchy–Schwarz} inequality:
    \begin{align}
        \left[\int_{K} \left(f(x)-f_{W}(x)\right)\left(f_{W}(x)-f_{CUR}(x)\right) d\mu \right]^2
        &\le \int_{K} \left(f(x)-f_{W}(x)\right)^2 d\mu \int_{K} \left(f_{W}(x)-f_{CUR}(x)\right)^2 d\mu \nonumber \\
        &\le \epsilon^2 \cdot \left(L^2 \|W_2\|_2^2 \|K\|_2^2 ((\eta_p + \eta_q)\sigma_{r+1})^2\right) \nonumber.
    \end{align}


    Combining all three parts \eqref{eq:proof-1}–\eqref{eq:proof-3}, we have:
    \[
        \| f - f_{CUR} \|_2^2 \le \epsilon^2 
        + L^2 \|W_2\|_2^2 \|K\|_2^2 ((\eta_p+\eta_q)\sigma_{r+1})^2 
        + 2 \epsilon L \|W_2\|_2 \|K\|_2 (\eta_p+\eta_q)\sigma_{r+1}.
    \]
    This can be expressed as:
    \[
        \| f - f_{CUR} \|_2^2 \le \left(\epsilon + L\|W_2\|_2\|K\|_2(\eta_p + \eta_q)\sigma_{r+1}\right)^2.
    \]

    To achieve \(\| f - f_{CUR} \|_2^2 \le (\epsilon + \delta)^2\), it suffices to set:
    \[
        L \|W_2\|_2 \|K\|_2 (\eta_p + \eta_q) \sigma_{r+1} \le \delta.
    \]
    Rearranging this inequality:
    \[
        \sigma_{r+1} \le \frac{\delta}{L (\eta_p + \eta_q)} \left( \| W_2 \|_2 \| K \|_2 \right)^{-1}.
    \]
\end{proof}


\subsection{Bounding MSE Using the Frobenius Norm} \label{appendix:proof2}

By Theorem~\ref{theorem:loss}, we show that the Frobenius norm-based loss \( \mathcal{L}(U) \) upper-bounds the Mean Squared Error (MSE).
\begin{proof}
Using Equation~\ref{eq:fc}, the MSE between the original model \(f_W\) and its approximated version \(f_{CUR}\) is defined as:
\begin{align}
    \text{MSE}(X)
    &= \frac{1}{b}\|f_{W}(X) - f_{CUR}(X)\|_F^2 \nonumber \\
    &= \frac{1}{b}\|\gamma(XW)W_2 - \gamma(XCUR)W_2 \|_F^2 \nonumber .
\end{align}
By extending the Lipschitz continuity from Equation~\ref{eq:lipschitz} to matrices, and so using the Frobenius norm:
\[
\|\gamma(XW) - \gamma(XCUR)\|_F \leq L\|X(W - CUR)\|_F.
\]
Further, since matrix multiplication is submultiplicative with respect to the Frobenius norm, it follows that:
\[
    \text{MSE}(X)
    = \frac{1}{b}\|\gamma(XW)W_2 - \gamma(XCUR)W_2 \|_F^2
    \leq \frac{1}{b} L^2 \|X\|_F^2 \|W_2\|_F^2 \|W - CUR\|_F^2.
\]
Rewriting the result using \(\mathcal{L}(U) = \|W - CUR\|_F^2\), we obtain:
\[
\text{MSE}(X) \leq \frac{1}{b} L^2 \|X\|_F^2 \|W_2\|_F^2 \mathcal{L}(U).
\]
Hence, minimizing \(\mathcal{L}(U)\) upper-bounds the MSE, offering an alternative for optimization.
\end{proof}


\subsection{Implicit Regularization in Healing} \label{appendix:proof3}

Theorem~\ref{theorem:direction} implies that the gradient of \(\mathcal{L}(U)\) cannot freely explore all directions in \(\mathbb{R}^{r \times r}\).
Instead, it is restricted to the subspace determined solely by the fixed matrices \(C\) and \(R\).

\begin{proof}
Given \( \mathcal{L}(U) = \| W - C U R \|_F^2 \), by the definition of the Frobenius norm, we have:
\[
    \mathcal{L}(U) = \| W - C U R \|_F^2 = \mathrm{trace}((W - C U R)^\top(W - C U R)).
\]

To find \(\nabla_{U} \mathcal{L}(U)\), we differentiate with respect to \(U\):
\begin{align}
    \nabla_{U} \mathcal{L}(U)
    &= \frac{\partial}{\partial U} \mathcal{L}(U) \nonumber \\
    &= \frac{\partial}{\partial U} \| W - C U R \|_F^2 \nonumber \\
    &= \frac{\partial}{\partial U} \left[ \mathrm{trace}((W - C U R)^\top(W - C U R)) \right] \nonumber \\
    &= \frac{\partial}{\partial U} \left[ \mathrm{trace}(W^\top W) \right] -2 \frac{\partial}{\partial U} \left[ \mathrm{trace}(W^\top CUR) \right] + \frac{\partial}{\partial U} \left[ \mathrm{trace}((CUR)^\top (CUR))\right] \nonumber \\
    &= 0 -2 C^\top W R^\top + 2 C^\top (CUR) R^\top \nonumber \\
    &= 2 C^\top (C U R - W) R^\top \nonumber.
\end{align}
Let \(M = (C U R - W) \in \mathbb{R}^{m \times n}\).
We can rewrite:
\begin{equation}
    \nabla_{U} \mathcal{L}(U) = 2 C^\top M R^\top.
    \label{eq:nabla-U}
\end{equation}

We now focus on the set:
\[
S = \{ C^\top M R^\top\}.
\]
To show that \(S\) is a subspace of \(\mathbb{R}^{r \times r}\), we verify conditions: zero vector in \(S\), closed under addition and scalar multiplication.

\begin{enumerate}
    \item
    Consider the zero matrix \(0_{m \times n}\), then \(C^\top (0_{m \times n}) R^\top = 0_{r \times r}\) is the zero element of \(\mathbb{R}^{r \times r}\). Thus, \(0_{r \times r} \in S\).

    \item
    Suppose \(A_1, A_2 \in S\). By definition, there exist \(M_1, M_2 \in \mathbb{R}^{m \times n}\) such that:
    \[
        A_1 = C^\top M_1 R^\top, \quad A_2 = C^\top M_2 R^\top.
    \]
    Consider their sum:
    \[
        A_1 + A_2 = C^\top M_1 R^\top + C^\top M_2 R^\top = C^\top (M_1 + M_2) R^\top.
    \]
    Since \(M_1 + M_2 \in \mathbb{R}^{m \times n}\), we have \( A_1 + A_2 \in S \).

    \item 
    Let \(A \in S\) and let \(\alpha \in \mathbb{R}\). There exists \(M \in \mathbb{R}^{m \times n}\) such that:
    \[
        A = C^\top M R^\top.
    \]
    Then:
    \[
        \alpha A = \alpha (C^\top M R^\top) = C^\top (\alpha M) R^\top.
    \]
    Since \(\alpha M \in \mathbb{R}^{m \times n}\), we have \(\alpha A \in S\).
\end{enumerate}
Hence, \(S\) contains the zero matrix, is closed under addition and scalar multiplication, \(S\) is indeed a subspace of \(\mathbb{R}^{r \times r}\).

By Equation~\ref{eq:nabla-U}, since \(\nabla_{U} \mathcal{L}(U) = 2 C^\top M R^\top\) for some \(M\), it follows that:
\[
    \nabla_{U} \mathcal{L}(U) \in S, \quad \text{i.e.,} \quad \nabla_{U} \mathcal{L}(U) \in \{ C^\top M R^\top\}.
\]
\end{proof}

\section{Hyperparameters} \label{appendix:hyperparameters}


\begin{itemize}
    \item \textbf{Batch Sizes}:
    During calibration, a single forward pass is performed using a default batch size of \(1\). For healing, the batch size is set to \(16\).
    
    \item \textbf{Learning Rate and Optimizer}:
    Following the layer-pruning research~\cite{gromov2024unreasonable}, we set the healing learning rate to \(3\times10^{-4}\).
    Optimization is performed using AdamW~\cite{loshchilov2017decoupled}, and a cosine learning rate scheduler~\cite{loshchilov2016sgdr} is employed with \(100\) warmup steps.
    
    \item \textbf{Knowledge Distillation Parameters}:
    A knowledge distillation weighting factor of \(\alpha = 0.1\) is applied.
    This means the student model learns from the teacher model weighted at 90\%, while a cross-entropy loss from the C4 dataset ground-truth is weighted at 10\%.
    The temperature parameter is set to \(T=10\) to facilitate effective distillation.
    
    \item \textbf{UUID Task}:
    We created a random UUID-to-UUID mapping task, presented in the form:
    \begin{center}
    \texttt{Given this UUID: <input\_uuid>{\textbackslash n}The corresponding UUID is: <output\_uuid>}
    \end{center}
    
    \item \textbf{LoRA and MoRA Settings}:
    For LoRA~\cite{hu2021lora}, we use \(\alpha=16\) and a dropout rate of \(0.1\). For MoRA~\cite{jiang2024mora}, we adopt the RoPE variant with a dropout rate of \(0.1\) as well.
\end{itemize}

\section{Comparisons} \label{appendix:comparison}

\subsection{Weight Selection} \label{appendix:weight-selection}

\begin{table*}[!th]
\footnotesize
\centering
\caption{
Performance comparison by target weights
(\textit{time} (s) above, \textit{parameters} middle, and \textit{size reduction} (GiB) below).
}
\resizebox{\textwidth}{!}{%
\begin{tabular}{c*{15}{|r}}
\hline
\multirow{2}{*}{\textbf{Model}} & \multicolumn{15}{c}{\textbf{Number of Compressed Layers}} \\ \cline{2-16}
& \multicolumn{1}{c|}{\textbf{2}} & \multicolumn{1}{c|}{\textbf{4}} & \multicolumn{1}{c|}{\textbf{6}} & \multicolumn{1}{c|}{\textbf{8}} & \multicolumn{1}{c|}{\textbf{10}} & \multicolumn{1}{c|}{\textbf{12}} & \multicolumn{1}{c|}{\textbf{14}} & \multicolumn{1}{c|}{\textbf{16}} & \multicolumn{1}{c|}{\textbf{18}} & \multicolumn{1}{c|}{\textbf{20}} & \multicolumn{1}{c|}{\textbf{22}} & \multicolumn{1}{c|}{\textbf{24}} & \multicolumn{1}{c|}{\textbf{26}} & \multicolumn{1}{c|}{\textbf{28}} & \multicolumn{1}{c}{\textbf{30}} \\ \hline \hline
\rowcolor{gray!30}
& 33.62 & 59.14 & 85.94 & 104.95 & 129.45 & 154.75 & 177.59 & 204.40 & 228.16 & 249.32 & 280.85 & 300.54 & 325.90 & 358.14 & 376.29 \\
\rowcolor{gray!30}
& 7.89B & 7.74B & 7.60B & 7.46B & 7.32B & 7.17B & 7.03B & 6.89B & 6.75B & 6.60B & 6.46B & 6.32B & 6.17B & 6.03B & 5.89B \\
\rowcolor{gray!30}
\multirow{-3}{*}{\textbf{All}}  
& $\blacktriangledown$0.53 & $\blacktriangledown$1.06 & $\blacktriangledown$1.60 & $\blacktriangledown$2.13 & $\blacktriangledown$2.66 & $\blacktriangledown$3.19 & $\blacktriangledown$3.72 & $\blacktriangledown$4.25 & $\blacktriangledown$4.79 & $\blacktriangledown$5.32 & $\blacktriangledown$5.85 & $\blacktriangledown$6.38 & $\blacktriangledown$6.91 & $\blacktriangledown$7.44 & $\blacktriangledown$7.98 \\ \hline
\multirow{3}{*}{\textbf{\(\{W^\text{Gate}\}\)}}
& 24.82 & 38.08 & 53.13 & 68.02 & 82.03 & 94.56 & 109.73 & 123.61 & 140.32 & 151.75 & 168.23 & 184.09 & 196.15 & 213.06 & 228.60 \\
& 7.92B & 7.81B & 7.71B & 7.60B & 7.49B & 7.38B & 7.28B & 7.17B & 7.06B & 6.95B & 6.84B & 6.74B & 6.63B & 6.52B & 6.41B \\
& $\blacktriangledown$0.40 & $\blacktriangledown$0.80 & $\blacktriangledown$1.21 & $\blacktriangledown$1.61 & $\blacktriangledown$2.01 & $\blacktriangledown$2.41 & $\blacktriangledown$2.81 & $\blacktriangledown$3.21 & $\blacktriangledown$3.62 & $\blacktriangledown$4.02 & $\blacktriangledown$4.42 & $\blacktriangledown$4.82 & $\blacktriangledown$5.22 & $\blacktriangledown$5.63 & $\blacktriangledown$6.03 \\ \hline
\multirow{3}{*}{\textbf{\(\{W^Q, W^K\}\)}}
& 19.69 & 29.50 & 38.65 & 48.33 & 57.67 & 67.93 & 78.74 & 87.74 & 97.72 & 105.88 & 117.25 & 126.65 & 135.79 & 148.53 & 157.17 \\
& 8.00B & 7.96B & 7.93B & 7.89B & 7.86B & 7.82B & 7.79B & 7.75B & 7.72B & 7.68B & 7.65B & 7.61B & 7.58B & 7.54B & 7.51B \\
& $\blacktriangledown$0.13 & $\blacktriangledown$0.26 & $\blacktriangledown$0.39 & $\blacktriangledown$0.52 & $\blacktriangledown$0.65 & $\blacktriangledown$0.78 & $\blacktriangledown$0.91 & $\blacktriangledown$1.04 & $\blacktriangledown$1.17 & $\blacktriangledown$1.30 & $\blacktriangledown$1.43 & $\blacktriangledown$1.56 & $\blacktriangledown$1.69 & $\blacktriangledown$1.82 & $\blacktriangledown$1.95 \\ \hline
\multirow{3}{*}{\textbf{\(\{W^Q, W^\text{Gate}\}\)}}
& 31.27 & 53.11 & 72.34 & 95.88 & 115.82 & 135.72 & 159.85 & 177.67 & 198.51 & 221.92 & 245.56 & 269.73 & 285.56 & 307.32 & 334.59 \\
& 7.89B & 7.76B & 7.62B & 7.48B & 7.34B & 7.21B & 7.07B & 6.93B & 6.80B & 6.66B & 6.52B & 6.39B & 6.25B & 6.11B & 5.97B \\
& $\blacktriangledown$0.51 & $\blacktriangledown$1.02 & $\blacktriangledown$1.53 & $\blacktriangledown$2.04 & $\blacktriangledown$2.55 & $\blacktriangledown$3.06 & $\blacktriangledown$3.58 & $\blacktriangledown$4.09 & $\blacktriangledown$4.60 & $\blacktriangledown$5.11 & $\blacktriangledown$5.62 & $\blacktriangledown$6.13 & $\blacktriangledown$6.64 & $\blacktriangledown$7.15 & $\blacktriangledown$7.66 \\ \hline
\multirow{3}{*}{\textbf{\(\{W^K, W^\text{Gate}\}\)}}
& 27.91 & 44.80 & 61.61 & 79.77 & 96.67 & 114.76 & 131.83 & 150.66 & 167.62 & 185.55 & 201.98 & 218.80 & 236.94 & 258.11 & 271.78 \\
& 7.92B & 7.80B & 7.69B & 7.58B & 7.46B & 7.35B & 7.24B & 7.12B & 7.01B & 6.90B & 6.78B & 6.67B & 6.55B & 6.44B & 6.33B \\
& $\blacktriangledown$0.42 & $\blacktriangledown$0.85 & $\blacktriangledown$1.27 & $\blacktriangledown$1.69 & $\blacktriangledown$2.11 & $\blacktriangledown$2.54 & $\blacktriangledown$2.96 & $\blacktriangledown$3.38 & $\blacktriangledown$3.81 & $\blacktriangledown$4.23 & $\blacktriangledown$4.65 & $\blacktriangledown$5.07 & $\blacktriangledown$5.50 & $\blacktriangledown$5.92 & $\blacktriangledown$6.34 \\ \hline
\end{tabular}%
}
\label{tab:weights}
\end{table*}

\input{texts/C_1_2_graph}

We evaluated various configurations for applying CUR decomposition to specific weight matrices, each yielding different trade-offs between model size reduction and performance.
Since our approach targets weight matrices before the activation function (to leverage the \(L\)-Lipschitz continuity assumption for error bounds), potential candidates include \( W^Q \) and \( W^K \) in the Multi-Head Attention layer (MHA), as well as \( W^{\text{Gate}} \) in the Feed-Forward Network (FFN).
\begin{itemize}
    \item \textbf{All Matrices (\( W^Q, W^K, W^{\text{Gate}} \))}: 
    Compressing all three produced acceptable performance degradation with the greatest model size reduction.
    We adopt this setting as our default.

    \item \textbf{FFN-only (\( W^{\text{Gate}} \))}:
    Targeting solely the FFN gate achieved strong performance (particularly on MMLU) and offered substantial size reductions.
    This option is suitable if a slightly lower compression rate is acceptable.

    \item \textbf{MHA-only (\( W^Q, W^K \))}:
    Applying CUR decomposition to just \( W^Q \) and \( W^K \) generally produced the best overall performance; however, size reduction was limited since the large FFN remained uncompressed.

    \item \textbf{\( W^Q \) and \( W^{\text{Gate}} \)}:
    Decomposing these weights demonstrated acceptable performance but did not outperform the \textbf{all matrices} option, making it somewhat less effective in terms of size reduction.

    \item \textbf{\( W^K \) and \( W^{\text{Gate}} \)}:
    This combination performed comparably to the \textbf{FFN-only} approach but achieved slightly greater size reduction. It is a viable alternative when one can afford a minor trade-off in performance.
\end{itemize}

Table~\ref{tab:weights} and Figure~\ref{fig:weight-selection} illustrate these size-performance trade-offs, based on Llama3.1-8B.
For C4 and WikiText2, we report perplexity (lower is better), whereas BoolQ and MMLU are measured by accuracy (higher is better). 
\subsection{Rank} \label{appendix:rank}

We also explore different maximum rank values, \(r_{\text{max}} \in \{128, 256, 512\}\).
Since our experiments focus on LLMs, the weight matrices typically have wide dimensions, so the rank \(r\) selection based on Equation~\ref{eq:r} is consistently constrained by the upper bound \(r_{\text{max}}\) (i.e., \(r \gets r_{\text{max}}\)).
Hence, changes to the maximum rank value \(r_{\text{max}}\) significantly influence performance.

As shown in Table~\ref{tab:ranks} and Figure~\ref{fig:ranks}, increasing \(r_{\text{max}}\) (e.g., from \(256\) to \(512\)) improves task performance but reduces compression efficiency and increases CURing computation time.
In contrast, \(r_{\text{max}}=128\) yields faster processing and greater compression but exhibits a more pronounced performance drop, especially on BoolQ.

\begin{table*}[!ht]
\footnotesize
\centering
\caption{
Performance comparison for different \(r_{\text{max}}\) settings across varying numbers of compressed layers (\textit{time} in seconds above, number of \textit{parameters} middle, and \textit{size reduction} in GiB below).
}
\resizebox{\textwidth}{!}{%
\begin{tabular}{c*{15}{|r}}
\hline
\textbf{Model}
                & \multicolumn{15}{c}{\textbf{Number of Compressed Layers}} \\ \cline{2-16}
\textbf{(Rank)}
                & \multicolumn{1}{c|}{\textbf{2}}   & \multicolumn{1}{c|}{\textbf{4}}   & \multicolumn{1}{c|}{\textbf{6}}   & \multicolumn{1}{c|}{\textbf{8}}   & \multicolumn{1}{c|}{\textbf{10}}  & \multicolumn{1}{c|}{\textbf{12}}  & \multicolumn{1}{c|}{\textbf{14}}  & \multicolumn{1}{c|}{\textbf{16}}  & \multicolumn{1}{c|}{\textbf{18}}  & \multicolumn{1}{c|}{\textbf{20}}  & \multicolumn{1}{c|}{\textbf{22}}  & \multicolumn{1}{c|}{\textbf{24}}  & \multicolumn{1}{c|}{\textbf{26}}  & \multicolumn{1}{c|}{\textbf{28}}  & \multicolumn{1}{c}{\textbf{30}}  \\ \hline \hline
                & 26.29 & 39.03 & 53.63 & 67.99 & 82.20 & 95.69 & 110.73 & 126.02 & 142.06 & 155.23 & 168.72 & 184.08 & 201.40 & 213.83 & 226.35 \\
                & 7.88B & 7.73B & 7.58B & 7.43B & 7.27B & 7.12B & 6.97B & 6.82B & 6.67B & 6.52B & 6.37B & 6.22B & 6.07B & 5.91B & 5.76B \\
\multirow{-3}{*}{\shortstack{\textbf{Llama3.1}\\\textbf{(128)}}}
                & $\blacktriangledown$0.56 & $\blacktriangledown$1.13 & $\blacktriangledown$1.69 & $\blacktriangledown$2.25 & $\blacktriangledown$2.82 & $\blacktriangledown$3.38 & $\blacktriangledown$3.94 & $\blacktriangledown$4.50 & $\blacktriangledown$5.07 & $\blacktriangledown$5.63 & $\blacktriangledown$6.19 & $\blacktriangledown$6.76 & $\blacktriangledown$7.32 & $\blacktriangledown$7.88 & $\blacktriangledown$8.45 \\ \hline
\rowcolor{gray!30}
                & 33.62 & 59.14 & 85.94 & 104.95 & 129.45 & 154.75 & 177.59 & 204.40 & 228.16 & 249.32 & 280.85 & 300.54 & 325.90 & 358.14 & 376.29 \\ 
\rowcolor{gray!30}
                & 7.89B & 7.74B & 7.60B & 7.46B & 7.32B & 7.17B & 7.03B & 6.89B & 6.75B & 6.60B & 6.46B & 6.32B & 6.17B & 6.03B & 5.89B \\
\rowcolor{gray!30}
\multirow{-3}{*}{\shortstack{\textbf{Llama3.1}\\\textbf{(256)}}}
                & $\blacktriangledown$0.53 & $\blacktriangledown$1.06 & $\blacktriangledown$1.60 & $\blacktriangledown$2.13 & $\blacktriangledown$2.66 & $\blacktriangledown$3.19 & $\blacktriangledown$3.72 & $\blacktriangledown$4.25 & $\blacktriangledown$4.79 & $\blacktriangledown$5.32 & $\blacktriangledown$5.85 & $\blacktriangledown$6.38 & $\blacktriangledown$6.91 & $\blacktriangledown$7.44 & $\blacktriangledown$7.98 \\ \hline
                & 54.05 & 97.85 & 142.38 & 185.40 & 226.54 & 269.22 & 322.92 & 364.50 & 405.71 & 453.12 & 504.17 & 545.39 & 597.89 & 635.62 & 687.00 \\
                & 7.90B & 7.78B & 7.65B & 7.53B & 7.40B & 7.28B & 7.15B & 7.03B & 6.90B & 6.78B & 6.65B & 6.53B & 6.40B & 6.28B & 6.15B \\                
\multirow{-3}{*}{\shortstack{\textbf{Llama3.1}\\\textbf{(512)}}}
                & $\blacktriangledown$0.47 & $\blacktriangledown$0.93 & $\blacktriangledown$1.40 & $\blacktriangledown$1.87 & $\blacktriangledown$2.33 & $\blacktriangledown$2.80 & $\blacktriangledown$3.27 & $\blacktriangledown$3.73 & $\blacktriangledown$4.20 & $\blacktriangledown$4.67 & $\blacktriangledown$5.13 & $\blacktriangledown$5.60 & $\blacktriangledown$6.07 & $\blacktriangledown$6.54 & $\blacktriangledown$7.00 \\ \hline
\end{tabular}%
}
\label{tab:ranks}
\end{table*}

\input{texts/5_2_3_comp_graph_2}

\input{texts/C_3_1_calibration_graph}
\subsection{Calibration} \label{appendix:calibration}

To obtain activations and measure angular distances, we perform a single forward pass for calibration.
As shown in Figure~\ref{fig:calibration}, increasing the calibration set size from \(128\) to \(1024\) shows negligible performance improvement, although slight differences may arise in specific cases (e.g., BoolQ at the 9-layer compression point).
However, calibration time scales linearly with the dataset size.
Hence, we use \(128\) examples by default.

\section{Ablation Analysis} \label{appendix:ablation}

\subsection{Layer Selection}

\begin{table*}[!ht]
\footnotesize
\centering
\caption{Per-layer angular distance, sorted in ascending order, using the 128 C4 calibration dataset.}
\resizebox{\textwidth}{!}{%
\begin{tabular}{r*{14}{|r}}
\hline
\multicolumn{15}{c}{\textbf{Layer \(n\) (sorted by the angular distance between layer \(n\) and layer \(n-1\))}} \\ \hline \hline
\multicolumn{1}{c|}{\cellcolor{gray!30}\textbf{25}}   & \multicolumn{1}{c|}{\cellcolor{gray!30}\textbf{26}}   & \multicolumn{1}{c|}{\cellcolor{gray!30}\textbf{27}}   & \multicolumn{1}{c|}{\cellcolor{gray!30}\textbf{24}}   & \multicolumn{1}{c|}{\cellcolor{gray!30}\textbf{28}}  & \multicolumn{1}{c|}{\cellcolor{gray!30}\textbf{23}}  & \multicolumn{1}{c|}{\cellcolor{gray!30}\textbf{22}}  & \multicolumn{1}{c|}{\cellcolor{gray!30}\textbf{29}}  & \multicolumn{1}{c|}{\cellcolor{gray!30}\textbf{20}}  & \multicolumn{1}{c|}{\cellcolor{gray!30}\textbf{21}}  & \multicolumn{1}{c|}{\textbf{19}}  & \multicolumn{1}{c|}{\textbf{18}}  & \multicolumn{1}{c|}{\textbf{17}}  & \multicolumn{1}{c|}{\textbf{30}}  & \multicolumn{1}{c}{\textbf{16}}  \\ \hline
\cellcolor{gray!30}0.0868 & \cellcolor{gray!30}0.0876 & \cellcolor{gray!30}0.0926 & \cellcolor{gray!30}0.0927 & \cellcolor{gray!30}0.0967 & \cellcolor{gray!30}0.0977 & \cellcolor{gray!30}0.1030 & \cellcolor{gray!30}0.1112 & \cellcolor{gray!30}0.1150 & \cellcolor{gray!30}0.1157 & 0.1199 & 0.1333 & 0.1522 & 0.1555 & 0.1616 \\ \hline \hline
\multicolumn{1}{c|}{\textbf{11}}   & \multicolumn{1}{c|}{\textbf{10}}   & \multicolumn{1}{c|}{\textbf{13}}   & \multicolumn{1}{c|}{\textbf{14}}   & \multicolumn{1}{c|}{\textbf{15}}  & \multicolumn{1}{c|}{\textbf{12}}  & \multicolumn{1}{c|}{\textbf{9}}  & \multicolumn{1}{c|}{\textbf{8}}  & \multicolumn{1}{c|}{\textbf{7}}  & \multicolumn{1}{c|}{\textbf{6}}  & \multicolumn{1}{c|}{\textbf{3}}  & \multicolumn{1}{c|}{\textbf{2}}  & \multicolumn{1}{c|}{\textbf{5}}  & \multicolumn{1}{c|}{\textbf{4}}  & \multicolumn{1}{c}{\textbf{1}}  \\ \hline
0.1685 & 0.1694 & 0.1717 & 0.1727 & 0.1749 & 0.1767 & 0.1827 & 0.1855 & 0.1972 & 0.2003 & 0.2053 & 0.2058 & 0.2099 & 0.2147 & 0.2254 \\ \hline
\end{tabular}%
}
\label{tab:angular}
\end{table*}

\input{texts/5_5_1_dist_graph}

Several studies on decoder-only LLM architectures suggest that the deeper part of the decoder contributes relatively little to changing the model’s output~\cite{gromov2024unreasonable, gloeckle2024better}.
In other words, these final layers do not substantially alter the intermediate representations they receive from preceding layers.
We investigate whether simply pruning these last few layers is sufficient or if our angular-distance-based approach provides benefits.

\begin{itemize}
    \item
    \textbf{Angular Distance}:
    Table~\ref{tab:angular} presents \textit{angular distances} between each pair of consecutive layers in Llama3.1-8B, measured on 128 examples from the C4 dataset.
    For instance, the value \(0.0868\) for layer \(25\) represents the angular distance between layer \(25\) and layer \(24\).
    As we discussed in Section~\ref{section:layer-selection}, the first and last layers are excluded from compression, so leaving 30 layers.
    These are sorted in ascending order by their distance; therefore, layers at the beginning of Table~\ref{tab:angular} are the ones deemed most similar to their preceding layers.
    We target these layers first for compression.
    %
    %

    \item
    \textbf{Last Layers}:
    By contrast, the \textit{last-\(N\)-layers} approach simply selects the last \(N\) layers for compression, excluding the model’s first and final layers.
\end{itemize}

Figure~\ref{fig:ablation-dist} compares these two layer-selection strategies, based on Llama3.1-8B.
On the C4 dataset, where the difference is minor but consistently in favor of angular distance.
As compression increases (more layers), accuracy drops more quickly for the last-\(N\)-layers method.
However, both methods consistently outperform a random selection of layers.
The last-\(N\)-layers approach shows performance that is not significantly behind the angular-distance method overall.

In practice, since activation collection (calibration) is already required for WANDA, leveraging these activations to compute angular distances brings additional performance gains at minimal cost.
Hence, while selecting last layers is a viable alternative, the angular-distance-based approach remains advantageous whenever calibration data is available.
As evidenced by the C4 perplexity results, diversifying the calibration dataset could potentially further enhance the benefits of angular distance selection across various tasks.


\subsection{WANDA \& DEIM}

\begin{table*}[!ht]
\footnotesize
\centering
\caption{
Comparison of per-layer \textit{Frobenius norms} (\(\Sigma\|W\|_F\) for original, \(\Sigma\| CUR \|_F\) for compressed) above, with \textit{differences} (\(\Sigma\| W - CUR \|_F\)) below in parentheses; layers sorted by ascending angular distance.
Smaller differences indicate a more accurate reconstruction.
}
\resizebox{0.84\textwidth}{!}{%
\begin{tabular}{c*{10}{|r}}
\hline
\multirow{2}{*}{\textbf{Model}} & \multicolumn{10}{c}{\textbf{Layers Sorted by Angular Distance}} \\ \cline{2-11}
                & \multicolumn{1}{c|}{\textbf{25}}   & \multicolumn{1}{c|}{\textbf{26}}   & \multicolumn{1}{c|}{\textbf{27}}   & \multicolumn{1}{c|}{\textbf{24}}   & \multicolumn{1}{c|}{\textbf{28}}  & \multicolumn{1}{c|}{\textbf{23}}  & \multicolumn{1}{c|}{\textbf{22}}  & \multicolumn{1}{c|}{\textbf{29}}  & \multicolumn{1}{c|}{\textbf{20}}  & \multicolumn{1}{c}{\textbf{21}}  \\ \hline \hline
\multirow{1}{*}{\textbf{Llama3.1}} 
                & 210.28 & 212.11 & 214.05 & 209.34 & 215.03 & 208.87 & 207.97 & 217.90 & 207.28 & 208.14 \\ \hline
\rowcolor{gray!30}
                & 156.10 & 157.74 & 160.65 & 154.93 & 162.56 & 153.10 & 152.28 & 166.21 & 150.76 & 151.75 \\ 
\rowcolor{gray!30}
\multirow{-2}{*}{{\textbf{CURing}}} 
                & (140.90) & (141.81) & (141.45) & (140.78) & (140.76) & (142.09) & (141.65) & (140.90) & (142.25) & (142.47) \\ \hline
%
%
\multirow{2}{*}{{\textbf{WANDA}}}
                & 155.16 & 157.30 & 160.00 & 154.53 & 162.17 & 152.62 & 151.88 & 165.75 & 150.37 & 151.53 \\ 
                & (141.93) & (142.30) & (142.18) & (141.22) & (141.21) & (142.61) & (142.07) & (141.44) & (142.66) & (142.69) \\ \hline
\multirow{2}{*}{{\textbf{DEIM}}}
                & 155.67 & 157.38 & 160.17 & 154.48 & 162.00 & 152.42 & 151.63 & 165.77 & 150.21 & 151.18 \\ 
                & (141.37) & (142.21) & (141.99) & (141.27) & (141.40) & (142.82) & (142.34) & (141.42) & (142.84) & (143.07) \\ \hline
\multirow{2}{*}{{\textbf{Weight}}}
                & 154.75 & 156.64 & 159.53 & 153.88 & 161.60 & 152.14 & 151.42 & 165.43 & 150.01 & 151.03 \\ 
                & (142.38) & (143.02) & (142.71) & (141.93) & (141.86) & (143.11) & (142.56) & (141.81) & (143.04) & (143.22) \\ \hline
\multirow{2}{*}{{\textbf{Random}}}
                & 153.95 & 155.53 & 158.19 & 152.86 & 160.79 & 151.22 & 150.20 & 164.21 & 148.36 & 149.28 \\ 
                & (143.24) & (144.23) & (144.19) & (143.02) & (142.78) & (144.09) & (143.85) & (143.22) & (144.76) & (145.05) \\ \hline
\end{tabular}%
}
\label{tab:wanda-deim}
\end{table*}

\input{texts/5_5_3_weight_graph}

We compare five approaches to selecting rows and columns for CUR decomposition, each differing in how it identifies the most important indices:
\begin{itemize}
    \item \textbf{CURing (WANDA + DEIM)}:
    Our proposed method, which combines both \textit{WANDA}~\cite{sun2023simple} and \textit{DEIM}~\cite{sorensen2016deim} for refined row/column selection.
    WANDA computes a fused importance matrix based on weight magnitudes and activation values from 128 examples of the C4 dataset.
    DEIM then uses singular value decomposition to iteratively select informative indices, removing redundancy as it proceeds.
    CURing applies DEIM to the WANDA importance matrix, thereby effectively choosing indices that capture both weight and activation information.

    \item \textbf{WANDA-only}:
    This method relies solely on WANDA’s fused information (weight magnitudes and activations), directly selecting rows and columns with the largest values.
    Unlike CURing, it does not perform any iterative singular-value-based screening.

    \item \textbf{DEIM-only}:
    This approach applies DEIM to the raw weight matrix, without activation information.
    It identifies the singular vectors with the largest contributions and removes overlaps among newly selected indices in an iterative fashion.

    \item \textbf{Weight}:
    A simpler baseline that considers only weight magnitudes.
    For each row or column, it calculates an \(ell_2\) norm divided by the Frobenius norm of the entire weight matrix; those with the largest scores are selected.

    \item \textbf{Random}:
    A purely random selection of row/column indices from the weight matrix to form \(C\) and \(R\).
\end{itemize}

Table~\ref{tab:wanda-deim} compares how closely each method approximates the original weights by reporting the Frobenius norm differences.
For the per-layer Frobenius norm, we summed the differences of all weights in the layer (\(\Sigma \| W - CUR \|_F\)).
We use Llama3.1-8B as the original model and compress 10 layers here.
Our CURing method, a combination of WANDA and DEIM, exhibits the smallest difference, indicating that it provides the best approximation of the original.
Although CUR decomposition itself is often sufficiently robust that even a random selection can appear close to the original, these seemingly small differences translate into pronounced variations in task performance.

Figure~\ref{fig:ablation-weight} shows the perplexity on C4 and WikiText2 and the accuracy on BoolQ and MMLU.
The Random method performs the worst overall, and considering only weight magnitudes (\textit{Weight Magnitude}) also results in poor performance.
By contrast, any use of WANDA (whether \textit{WANDA-only} or \textit{CURing}) yields stronger results on C4, likely due to integrating activation information derived from that dataset.
Among all approaches, \textit{CURing} shows the most stable performance as the number of compressed layers increases, outperforming alternatives on tasks like WikiText2 perplexity and BoolQ accuracy.

These findings underscore the benefits of combining WANDA’s activation-aware information with DEIM’s iterative, redundancy-reducing selection strategy, demonstrating the efficacy of our proposed approach.
%

\section{Activation Analysis} \label{appendix:activations}

\begin{table*}[!ht]
\footnotesize
\centering
\caption{Comparison of per-weight activation \textit{Frobenius Norms} above, with \textit{differences} (\(\|W - CUR\|_F\)) below.}
%
%
\resizebox{1.0\textwidth}{!}{%
\begin{tabular}{c*{15}{|r}}
\hline
\multirow{2}{*}{\textbf{Model}} & \multicolumn{3}{c|}{\textbf{Layer 25}} & \multicolumn{3}{c|}{\textbf{Layer 26}} & \multicolumn{3}{c|}{\textbf{Layer 27}} & \multicolumn{3}{c|}{\textbf{Layer 24}} & \multicolumn{3}{c}{\textbf{Layer 28}} \\ 
\cline{2-16}
& \textbf{\(W^Q\)} & \textbf{\(W^K\)} & \textbf{\(W^\text{Gate}\)} & \textbf{\(W^Q\)} & \textbf{\(W^K\)} & \textbf{\(W^\text{Gate}\)} & \textbf{\(W^Q\)} & \textbf{\(W^K\)} & \textbf{\(W^\text{Gate}\)} & \textbf{\(W^Q\)} & \textbf{\(W^K\)} & \textbf{\(W^\text{Gate}\)} & \textbf{\(W^Q\)} & \textbf{\(W^K\)} & \textbf{\(W^\text{Gate}\)} \\
\hline \hline
\textbf{Llama3.1}
    & 42.48 & 24.72 & 18.88 & 28.72 & 21.39 & 21.86 & 34.80 & 23.40 & 23.48 & 36.34 & 22.70 & 15.56 & 31.04 & 26.11 & 25.62 \\
\hline
\rowcolor{gray!30}
    & 51.83 & 30.83 & 21.89
    & 34.05 & 25.74 & 23.60
    & 43.71 & 28.43 & 25.06
    & 44.60 & 28.68 & 18.12
    & 38.54 & 31.66 & 29.04 \\
\rowcolor{gray!30}\multirow{-2}{*}{\textbf{CURing}}
    & (9.86) & (6.78) & (4.10)
    & (6.54) & (5.39) & (3.25)
    & (9.66) & (6.17) & (3.41)
    & (8.79) & (6.58) & (3.37)
    & (8.36) & (6.97) & (5.68) \\
\hline
\rowcolor{gray!30}\textbf{CURing}
    & 43.92 & 25.57 & 18.97
    & 29.52 & 22.17 & 21.76
    & 35.88 & 23.99 & 22.27
    & 36.94 & 23.18 & 16.02
    & 30.20 & 25.51 & 22.40 \\
\rowcolor{gray!30}\textbf{(Healed)}
    & (1.88) & (1.63) & (1.53)
    & (1.78) & (1.95) & (1.34)
    & (2.39) & (1.93) & (2.07)
    & (1.36) & (1.33) & (1.01)
    & (2.52) & (2.35) & (4.46) \\
\hline \hline
\multirow{2}{*}{\textbf{Model}} & \multicolumn{3}{c|}{\textbf{Layer 23}} & \multicolumn{3}{c|}{\textbf{Layer 22}} & \multicolumn{3}{c|}{\textbf{Layer 29}} & \multicolumn{3}{c|}{\textbf{Layer 20}} & \multicolumn{3}{c}{\textbf{Layer 21}} \\ 
\cline{2-16}
& \textbf{\(W^Q\)} & \textbf{\(W^K\)} & \textbf{\(W^\text{Gate}\)} & \textbf{\(W^Q\)} & \textbf{\(W^K\)} & \textbf{\(W^\text{Gate}\)} & \textbf{\(W^Q\)} & \textbf{\(W^K\)} & \textbf{\(W^\text{Gate}\)} & \textbf{\(W^Q\)} & \textbf{\(W^K\)} & \textbf{\(W^\text{Gate}\)} & \textbf{\(W^Q\)} & \textbf{\(W^K\)} & \textbf{\(W^\text{Gate}\)} \\
\hline \hline
\textbf{Llama3.1}
    & 28.32 & 24.68 & 15.78 & 36.37 & 24.11 & 15.59 & 23.24 & 30.38 & 24.70 & 27.24 & 21.98 & 16.60 & 39.18 & 24.82 & 16.48 \\
\hline
\rowcolor{gray!30}
    & 35.91 & 31.01 & 18.38
    & 44.86 & 30.10 & 18.30
    & 29.35 & 37.57 & 31.82
    & 27.24 & 21.98 & 20.60
    & 49.31 & 30.81 & 19.10 \\
\rowcolor{gray!30}\multirow{-2}{*}{\textbf{CURing}}
    & (7.99) & (6.80) & (3.20)
    & (8.81) & (6.50) & (3.14)
    & (8.27) & (8.70) & (8.32)
    & (0.00) & (0.00) & (4.60)
    & (10.32) & (6.87) & (3.08) \\
\hline
\rowcolor{gray!30}\textbf{CURing}
    & 26.90 & 23.57 & 15.12
    & 34.30 & 23.16 & 14.89
    & 22.01 & 29.63 & 21.09
    & 27.24 & 21.98 & 16.82
    & 36.65 & 23.52 & 15.23 \\
\rowcolor{gray!30}\textbf{(Healed)}
    & (1.91) & (1.70) & (1.09)
    & (2.36) & (1.54) & (1.00)
    & (2.90) & (2.69) & (4.88)
    & (0.00) & (0.00) & (0.30)
    & (2.66) & (1.66) & (1.39) \\
\hline
\end{tabular}%
}
\label{tab:activations}
\end{table*}

Table~\ref{tab:activations} compares the per-weight activation Frobenius norms of Llama3.1-8B (teacher), the \(10\)-layer CURing-compressed model (student), and the same student after healing with Knowledge Distillation (KD).
The activations are gathered using the C4 validation dataset.
The numbers in parentheses represent \(\|W - CUR\|_F\), capturing the difference between the original weights and the compressed ones.
Although these differences are initially present (if not very large), once KD is applied, they shrink considerably.
%
%
%
%
%
These observations underscore the interpretability benefits of CURing.
By preserving row and column subsets from the original model, CURing inherently retains much of the network’s characteristics.
The optional healing phase enhances this further, refining the student’s activations to more closely mirror the teacher’s.



\end{document}